\pdfoutput=1 
\documentclass{article}

\usepackage{PRIMEarxiv}

\usepackage[utf8]{inputenc} 
\usepackage[T1]{fontenc}    
\usepackage{amsmath,amssymb,amsfonts}
\usepackage[dvipsnames]{xcolor}
\usepackage{colortbl}
\usepackage{graphicx}
\usepackage{booktabs}
\usepackage{tabularx}
\usepackage{multirow}
\usepackage{makecell}
\usepackage[flushleft]{threeparttable}
\usepackage{pifont}
\usepackage{algorithm}
\usepackage{algpseudocode}
\usepackage{wrapfig} 
\usepackage[numbers,sort&compress]{natbib}
\usepackage{url}
\usepackage{nicefrac}
\usepackage{microtype}
\usepackage{placeins} 
\usepackage{fancyhdr}
\usepackage[bottom]{footmisc} 
\usepackage[colorlinks=true,linkcolor=NavyBlue,citecolor=NavyBlue,urlcolor=NavyBlue]{hyperref}
\graphicspath{{figs/}}

\algrenewcommand\algorithmicrequire{\textbf{Input:}}
\algrenewcommand\algorithmicensure{\textbf{Output:}}
\algrenewcommand{\algorithmiccomment}[1]{\hfill{\color{black!55}\(\triangleright\)~#1}}
\algnewcommand{\Gets}{\leftarrow}
\algnewcommand{\IfThen}[2]{\State \algorithmicif\ #1\ \algorithmicthen\ #2} 

\newcommand{\sys}{\textsc{LAST-V2X}}
\newcommand{\code}[1]{\texttt{#1}}
\newcommand{\Circled}[1]{\ding{\the\numexpr181+#1\relax}}

\definecolor{TabGroup}{gray}{0.93}
\definecolor{TabFull}{RGB}{232,240,250}
\definecolor{SevError}{RGB}{190,50,45}
\definecolor{SevWarn}{RGB}{214,140,20}
\definecolor{SevNote}{RGB}{110,118,128}
\newcommand{\sevtag}[2]{{\setlength{\fboxsep}{1.5pt}\colorbox{#1}{\textcolor{white}{\scriptsize\sffamily\bfseries\strut #2}}}}

\usepackage{tcolorbox}
\tcbuselibrary{skins}
\definecolor{PromptHead}{RGB}{61,90,128}
\definecolor{PromptBg}{RGB}{247,249,252}
\newtcolorbox{promptpanel}[2]{enhanced, colback=PromptBg, colframe=PromptHead, colbacktitle=PromptHead, coltitle=white, boxrule=0.4pt, arc=1.2pt, left=4pt, right=4pt, top=2pt, bottom=2.5pt, toptitle=1pt, bottomtitle=1pt, fonttitle=\sffamily\bfseries\footnotesize, fontupper=\footnotesize\raggedright, title={#1\hfill{\mdseries #2}}, before skip=3pt, after skip=3pt}
\newenvironment{pfields}{\begin{list}{}{\setlength{\leftmargin}{5.2em}\setlength{\labelwidth}{4.7em}\setlength{\labelsep}{0.5em}\setlength{\itemsep}{1.5pt}\setlength{\parsep}{0pt}\setlength{\topsep}{0.5pt}\setlength{\partopsep}{0pt}}\raggedright}{\end{list}}
\newcounter{prule}
\newenvironment{prules}{\begin{list}{\color{PromptHead}(\roman{prule})}{\usecounter{prule}\setlength{\leftmargin}{1.9em}\setlength{\labelwidth}{1.6em}\setlength{\labelsep}{0.3em}\setlength{\itemsep}{0.5pt}\setlength{\parsep}{0pt}\setlength{\topsep}{0pt}\setlength{\partopsep}{0pt}}}{\end{list}}
\newenvironment{pnotes}{\begin{list}{\color{PromptHead}\textbullet}{\setlength{\leftmargin}{0.9em}\setlength{\labelwidth}{0.6em}\setlength{\labelsep}{0.3em}\setlength{\itemsep}{0.5pt}\setlength{\parsep}{0pt}\setlength{\topsep}{0pt}\setlength{\partopsep}{0pt}}}{\end{list}}

\title{\textsc{LAST-V2X}: Language-Driven Adaptive Scenario Synthesis and Testing for Communication-Induced Failures in V2X Safety Applications}

\author{
  Keshu Wu$^{1,2}$, Hao Zhang$^{1}$, Rui Gan$^{2}$, Pei Li$^{3}$, Jiwan Jiang$^{1}$, Yang Zhou$^{1,*}$ \\[4pt]
  $^{1}$Texas A\&M University,
  $^{2}$University of Wisconsin-Madison,
  $^{3}$University of Wyoming \\
  [8pt]
}

\begin{document}
\maketitle
\vspace{-0.56in} 

\begin{abstract}
Vehicle-to-everything (V2X) communication enables connected vehicles to make safety-critical decisions, but delayed, missing, or incorrect information can compromise driving safety. Systematically testing such failures is challenging because traffic interactions, communication conditions, and application responses must be jointly configured to produce meaningful scenarios. This paper presents LAST-V2X, a language-driven, adaptive framework for automated scenario synthesis and safety testing of V2X applications. LAST-V2X leverages large language models (LLMs) to translate natural-language test descriptions into executable scenarios grounded in roadway geometry, V2X communication semantics, and vehicle behavior. The framework integrates physical traffic simulation with spatially aware communication modeling and message-level fault injection to evaluate how information degradation affects vehicle decisions. To establish whether an observed safety failure is caused by communication rather than the underlying traffic conditions, LAST-V2X employs matched executions under degraded and ground-truth information. Furthermore, an execution-guided refinement mechanism combines parameter search and LLM self-reflection to identify ineffective test configurations, improve scenario effectiveness, and produce minimal reproducible failure cases. Experiments across diverse CARLA environments and V2X safety applications demonstrate the framework's ability to generate reliable scenarios, distinguish communication-induced failures from unrelated safety violations, and substantially improve failure discovery through adaptive refinement. These results establish LAST-V2X as a systematic approach to language-driven, causally verified, and adaptive safety testing for connected vehicle systems.
\end{abstract}


\begin{figure}[!b]
  \centering
  \includegraphics[width=0.96\textwidth]{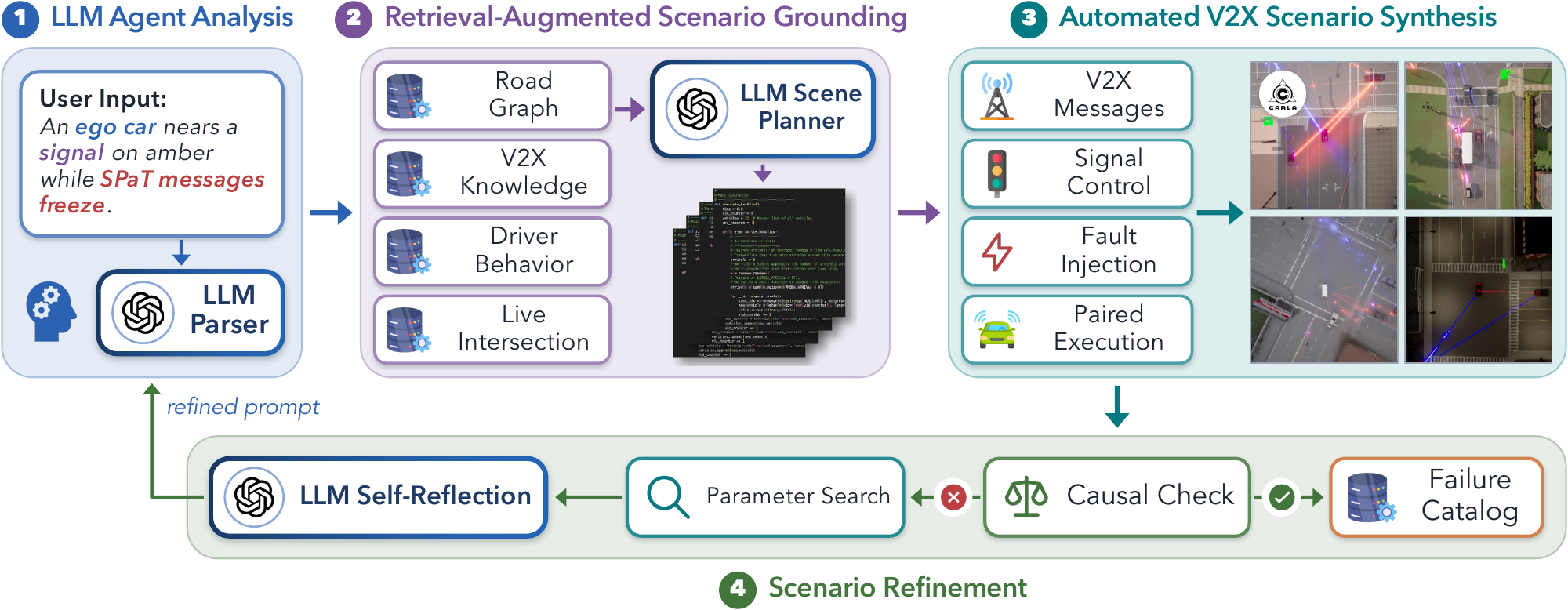}\\[2pt]
  {\small\textbf{Overview of LAST-V2X\protect\footnotemark.} LLM-driven synthesis of V2X test scenarios with causal checks and self-reflective refinement.\par}
\end{figure}
\footnotetext{Project page: \url{https://keshuw95.github.io/LAST-V2X/}}

\section{Introduction}
\label{sec:intro}

Vehicle-to-everything (V2X) communication extends a vehicle's awareness beyond onboard sensors \cite{YI2024100126,yang2026edge,li2026cats,gao2025airv2x,yu2025end}. Roadside units (RSUs) broadcast Signal Phase and Timing (SPaT) and intersection geometry (MAP) messages, while connected vehicles exchange Basic Safety Messages (BSMs). Applications such as Red-Light Violation Warning (RLVW), Intersection Movement Assist (IMA), and Left Turn Assist (LTA) use this information to support safety-critical decisions \cite{sae-j2735_2016,wu2023development,yu2023v2x}. This communication--driving coupling also creates a cyber--physical safety pathway: delayed, missing, or incorrect V2X information can alter vehicle decisions and ultimately compromise safety. Systematically testing such communication-induced failures is therefore increasingly important. Constructing these tests manually is difficult because a meaningful V2X scenario must coordinate roadway, traffic, vehicle behavior, message type, communication condition, and fault timing. These elements are strongly coupled: a delayed SPaT message, for example, matters only when it occurs at the right movement, phase transition, and decision time. The resulting scenario space is large and highly structured, and many seemingly reasonable combinations are infeasible or ineffective.

Beyond the difficulty of specifying coupled V2X scenarios, constructing them in high-fidelity simulators can itself require substantial effort to understand simulator APIs, scenario logic, traffic control, and communication interfaces. Large language models (LLMs) \cite{singh2025openai,wang2025generative} offer a natural bridge between engineers' high-level safety intents and the complex simulation logic required to instantiate executable V2X test scenarios. Transportation engineers can naturally express safety concerns in language. An LLM can translate such intent into structured requirements, identify the relevant safety application, connect semantic traffic roles with domain concepts, and retrieve the roadway, communication, and behavioral knowledge needed to instantiate the test \cite{wu2025v2x,you2024v2x,liu2025colmdriver,chiu2026v2v,gao2025langcoop}. Recent language-driven methods have demonstrated the potential of compiling natural-language descriptions into executable driving scenes \cite{ruan2025tts,zhang2024chatscene,fremont2023scenic,ding2024realgen,sheng2025talk2traffic,tu2025trafficcomposer,cai2026text2scenario,wu2026aurora,gao2025laser,tan2025promptable,elmaaroufi2024scenicnl}, a direction recently surveyed in~\cite{zhao2026survey}. This capability is particularly valuable for V2X, where a scenario depends not only on \emph{how vehicles move}, but also on \emph{what information they exchange and what each vehicle believes about the traffic state}. LLMs can therefore reduce both scenario-engineering effort and simulator-programming burden by orchestrating heterogeneous transportation, communication, and simulation knowledge.

Despite this promise, generating a plausible V2X scenario does not guarantee a meaningful safety test. A communication fault matters only if it changes the information used by the V2X application and, in turn, alters the vehicle's physical behavior or safety outcome. This cyber--physical effect can be difficult to establish because communication failures take different forms, including packet loss, latency, timestamp errors, decoding delays, frozen controller messages, position errors, and incorrect lane-to-signal-group associations \cite{li2023does,wu2025digital,liao2024digital,zhang2025virtual,zhang2024fabrication}. Some missing or delayed messages become stale and can be detected by the application, potentially triggering a safe fallback. In contrast, incorrect information may still appear fresh and therefore be trusted, leading the vehicle to act on a wrong traffic state. Safety-critical information may also be absent entirely, for example when a conflicting vehicle does not transmit V2X messages. Therefore, language-driven generation alone is insufficient: the generated scenario must explicitly model how communication faults affect application decisions and verify whether those effects actually propagate into physical safety consequences.

Existing safety-critical scenario-generation methods primarily create risk by modifying the \emph{physical world} \cite{dosovitskiy2017carla,li2022metadrive,feng2021intelligent,feng2023dense,wu2025ai2,wu2025hypergraph}: vehicle trajectories, behaviors, road conditions, or initial states are perturbed until an unsafe interaction emerges \cite{ding2020learning,wang2021advsim,zhang2023cat,xu2022safebench}. Recent language-driven approaches further allow such physical scenarios to be specified and generated from natural-language descriptions \cite{zhang2024chatscene,sheng2025talk2traffic,LI2026105759,tang2024legend,peng2026ldscene,mei2025llmattacker,mei2025seeking,gao2025words,wang2026learning,zhao2026sg,zhou2025decoupled,guo2024sovar}. V2X safety testing, however, poses a fundamentally different problem. The physical traffic scene may remain unchanged while communication causes the ego vehicle to perceive a different state of that scene. A signal may physically turn red while the received SPaT still indicates green; a conflicting vehicle may physically exist while its BSM is missing; or the roadway geometry may be correct while the received MAP associates the ego with the wrong movement. The safety-critical condition therefore arises not simply from \emph{what happens in the physical world}, but from a mismatch between the physical world and the information available to the vehicle.

V2X co-simulation and digital-twin platforms provide mechanisms for modeling message exchange, latency, packet loss, and communication-dependent vehicle control \cite{cai2023,zelenbaba2022wireless,liu2022application,maglogiannis2021experimental,wu2025digital,xu2023opencda,grimm2024cosimulate}. Automated testing of V2X systems has so far focused mainly on cooperative perception, for example by transforming shared sensor data under varied communication and weather conditions \cite{guo2024cootest}, rather than on message-level faults in closed-loop safety applications. However, communication simulation alone does not solve the testing problem. The physical interaction, message semantics, fault type, fault timing, and application response must be jointly constructed so that the information mismatch occurs precisely when it can affect a safety-critical decision. A generated test may also be executable yet ineffective because the selected fault is mistimed, too weak, or semantically inconsistent with the intended failure mechanism. Existing language-driven methods largely stop after scenario generation or, when they close the loop, refine only the physical adversarial behavior \cite{mei2025llmattacker}, while communication simulators typically assume that the scenario and fault configuration are already specified; neither provides a mechanism to use execution evidence to diagnose why a test failed and revise the underlying test intent. Moreover, observing a collision after injecting a communication fault is insufficient to establish a V2X-induced failure, because the same collision may occur even under healthy communication. Existing language-driven scenario generation, communication simulation, and safety falsification therefore address different parts of the problem, but do not provide an integrated, self-refining framework for constructing, diagnosing, refining, and causally validating failures that originate in the information layer and propagate into the physical world. This leads to the central question of this work: \emph{can natural-language V2X test intents be transformed into executable scenarios that systematically expose and causally verify communication-induced safety failures?}

\begin{figure*}[!t]
  \centering
  \includegraphics[width=0.98\textwidth]{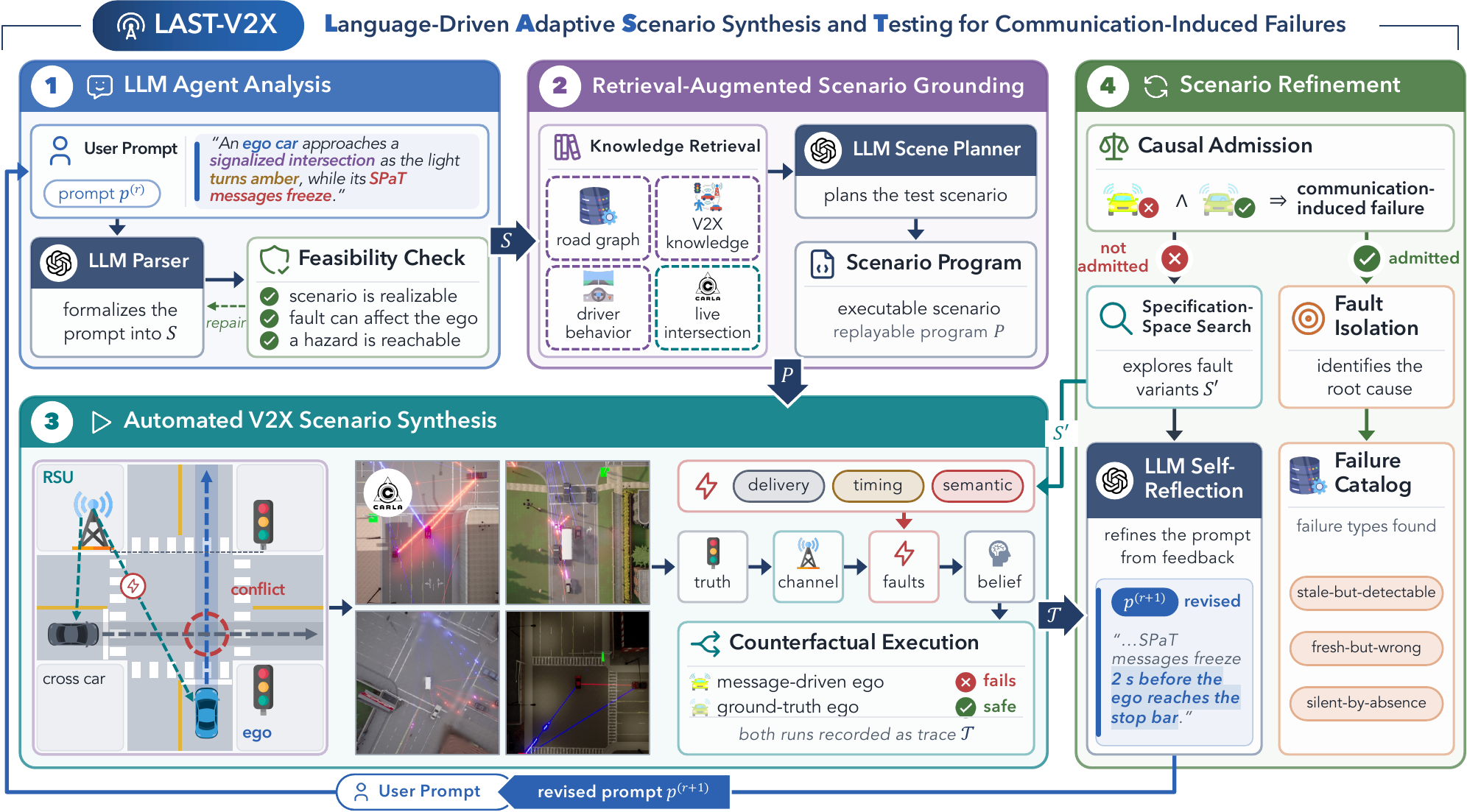}
\caption{Overview of \sys{}. \emph{LLM Agent Analysis} verifies the intent; \emph{Retrieval-Augmented Scenario Grounding} places it on a feasible junction; \emph{Automated V2X Scenario Synthesis} runs it in CARLA with faulted messages beside a ground-truth ego; \emph{Scenario Refinement} admits a failure only if the message-driven ego alone fails, else refines it.}
  \label{fig:pipeline}
\end{figure*}

To address this question, we develop \sys{}, an LLM-orchestrated framework for communication-aware V2X safety testing. \sys{} first converts natural-language intents into grounded scenario specifications, then executes coupled traffic--communication scenarios in CARLA under matched degraded and non-degraded communication conditions. Execution evidence is further used to refine ineffective scenarios through parameter search and LLM self-reflection, while successful failures are reduced to minimal reproducible cases.

The main contributions are:
{\setlength{\topsep}{2pt}\setlength{\partopsep}{0pt}
\begin{enumerate}\setlength{\itemsep}{1pt}\setlength{\parsep}{0pt}\setlength{\parskip}{0pt}

\item \textbf{Coupled physical--information scenario generation.} \sys{} jointly constructs the physical traffic scene and the V2X information perceived by the vehicle, with feasibility checks to reject ineffective tests before simulation.

\item \textbf{Causal verification of communication-induced safety failures.} Matched degraded and non-degraded executions determine whether V2X faults actually cause changes in physical safety outcomes.

\item \textbf{Execution-grounded adaptive refinement.} Parameter search corrects quantitative deficiencies, while LLM self-reflection revises semantically ineffective test intents using simulation evidence. Successful cases are further reduced to minimal reproducible failures.
\end{enumerate}}

The remainder of this paper is organized as follows. Section~\ref{sec:method} formulates the testing problem and introduces the four modules of \sys{}. Section~\ref{sec:experiments} presents the experimental evaluation, and Section~\ref{sec:conclusion} concludes the paper.

\section{Methodology}
\label{sec:method}

As illustrated in Fig.~\ref{fig:pipeline}, \sys{} contains four interdependent modules. \emph{LLM Agent Analysis} converts the user intent into a typed and feasibility-verified test specification. \emph{Retrieval-Augmented Scenario Grounding} binds the symbolic specification to roadway geometry, V2X domain knowledge, and executable agent behaviors. \emph{Automated V2X Scenario Synthesis} realizes the resulting program as a coupled physical--information simulation with spatial communication, semantic message faults, and paired ego rollouts. Finally, \emph{Scenario Refinement} uses execution evidence to search for, diagnose, and minimize communication-induced failures. We first formalize the problem solved by these four modules.

\subsection{Problem Description}
\label{sec:problem}

\paragraph{Scenario Specification}
The input to \sys{} is a short free-form natural-language prompt $p$ describing a traffic situation to be tested. The prompt may specify the environment, ego maneuver, surrounding vehicles, safety-critical interaction, and optionally a communication condition or fault. A compiler $\mathcal{C}$ transforms the prompt into a structured scenario specification
\begin{equation}
S
=
\mathcal{C}(p)
=
\big(
a,\,
E,\,
Q,\,
\mathcal{V},\,
u_{\mathrm{ego}},\,
C,\,
F
\big),
\label{eq:spec}
\end{equation}
where $a$ denotes the V2X safety application under test; $E$ describes environmental conditions; $Q$ specifies geometric and traffic-control requirements; $\mathcal{V}$ is the roster of surrounding agents; $u_{\mathrm{ego}}$ defines the ego maneuver and initial conditions; $C$ describes the nominal communication configuration; and $F=\{F_1,\ldots,F_m\}$ denotes the set of communication faults.

Each non-ego vehicle $i\in\mathcal{V}$ is characterized by its vehicle class, semantic role in the interaction, initial distance and speed, V2X equipage status, and behavioral character. The communication configuration includes message broadcast rates, communication parameters, V2X penetration, and the signal plan $\Psi$. Each fault specifies a message-level operator, its target information item, an event-anchored activation window, and its associated magnitude or parameters.

\paragraph{Physical and Perceived States}
Each scenario exercises one V2X application $a$, represented as a decision function
\begin{equation}
\pi_a(\beta,x)\rightarrow u,
\label{eq:decision}
\end{equation}
where $x$ is the ego physical state, $\beta$ is the information state reconstructed from received V2X messages, and $u$ is the resulting control action. Importantly, the application does not act directly on the simulator's ground-truth world state. Its behavior depends on what information has been sampled, communicated, possibly degraded, and reconstructed at the receiver.

We represent this transformation through the information chain
\begin{equation}
W
\;\xrightarrow{\;\Sigma\;}\;
\mathcal{M}
\;\xrightarrow{\;\Gamma\;}\;
\mathcal{M}'
\;\xrightarrow{\;\Phi_F\;}\;
\mathcal{M}''
\;\xrightarrow{\;\mathcal{B}\;}\;
\beta ,
\label{eq:chain}
\end{equation}
where $W(t)$ is the ground-truth world state, $\Sigma$ is message sampling, $\Gamma$ is communication-channel delivery, $\Phi_F$ is fault injection, and $\mathcal{B}$ is receiver-side state reconstruction. Thus, $\beta(t)$ is generally only an imperfect representation of $W(t)$.

A generated message is represented by
\begin{equation}
m_k=(t_k,\hat{t}_k,y_k),
\end{equation}
where $t_k$ is its true generation time, $\hat{t}_k$ is the timestamp reported by the message, and $y_k$ is its semantic payload. For information item $i$, the receiver retains the most recent accepted payload,
\begin{equation}
\beta_i(t)=y_{k^\ast,i},
\end{equation}
whose Age of Information is
\begin{equation}
\Delta_i(t)
=
t-\hat{t}_{k^\ast}.
\label{eq:aoi-def}
\end{equation}

This representation distinguishes three information-failure mechanisms used throughout the paper. In a \emph{stale-but-detectable} condition, loss or delay increases $\Delta_i(t)$ and can trigger a freshness-based fallback. In a \emph{fresh-but-wrong} condition, messages remain apparently current while their semantic content differs from the physical world. A \emph{silent-by-absence} condition occurs when safety-relevant information is never generated, for example because a conflicting vehicle is not V2X-equipped.

\paragraph{Causal Failure Objective}
A difference between $\beta(t)$ and $W(t)$ is not itself a safety failure, and an observed collision is not necessarily caused by communication. The objective of \sys{} is therefore to discover scenarios in which degraded V2X information changes the safety outcome.

For an executed scenario $S$, let $V_{\mathrm{msg}}(S)\in\{0,1\}$ indicate whether the message-driven ego violates the application-specific safety criterion, and let $V_{\mathrm{gt}}(S)\in\{0,1\}$ denote the corresponding outcome when the same application acts on ground-truth information. We define the causal-admission indicator
\begin{equation}
A(S)
=
\mathbb{I}
\left[
V_{\mathrm{msg}}(S)=1
\;\wedge\;
V_{\mathrm{gt}}(S)=0
\right].
\label{eq:admission}
\end{equation}
A scenario is therefore admitted as a communication-induced failure only when the message-driven ego fails while its matched ground-truth control remains safe.

The primary falsification objective is
\begin{equation}
\max_{S\in\mathcal{F}} A(S),
\label{eq:objective}
\end{equation}
where $\mathcal{F}$ denotes the set of executable specifications satisfying the schema, analytic feasibility conditions, roadway constraints, and applicable intent-preservation requirements.

Since $A(S)$ is binary and failures can be sparse, the search is guided before admission by a continuous failure margin $m(S)$,
\begin{equation}
\min_{S'\in\mathcal{N}(S)\cap\mathcal{F}} m(S'),
\label{eq:margin-objective}
\end{equation}
where $\mathcal{N}(S)$ denotes bounded variations of the current scenario. Smaller margins indicate scenarios closer to violating the target safety criterion and are therefore prioritized during refinement. Appendix~\ref{app:margin} defines $m(S)$ and the search settings.

Once an admitted failure has been identified, the objective changes from discovery to explanation. Let $F(S)$ denote its injected fault set. The minimal-witness problem is
\begin{equation}
F^\ast
=
\mathop{\arg\min}_{F'\subseteq F(S)}
|F'|
\qquad
\mathrm{s.t.}
\qquad
A(S[F'])=1,
\label{eq:minimal-witness}
\end{equation}
so the final artifact retains only the smallest fault configuration required to reproduce the communication-caused failure; for a silent-by-absence failure, $F(S)=\emptyset$ and the witness is the unfaulted scenario itself. At the benchmark level, multiple such witnesses are sought across different applications, fault classes, penetration levels, and conflict geometries rather than repeatedly maximizing the severity of a single mechanism.

The four modules below implement this process from language interpretation to causally verified failure discovery.

\subsection{Module I: LLM Agent Analysis}
\label{sec:parser}

The first module converts an open-ended natural-language request into a structured V2X test that is both syntactically valid and analytically capable of exercising the intended communication mechanism. It combines a schema-constrained LLM semantic parser with a deterministic feasibility verifier. The LLM determines what the prompt means; explicit rules and arithmetic determine whether that interpretation can produce a meaningful test.

\paragraph{Schema-Constrained Intent Parsing}

Let $\Lambda$ denote the language interpretation operator. Given prompt $p$, it produces an initial candidate specification
\begin{equation}
S^{(0)}=\Lambda(p).
\end{equation}
The parser uses structured JSON output constrained by the schema of Eq.~(\ref{eq:spec}). It identifies the application under test, environment and map requirements, ego and surrounding-agent attributes, communication configuration, and requested faults. When the prompt does not explicitly name the safety application, $\Lambda$ infers it from the traffic interaction. Explicitly specified communication conditions are preserved, whereas unspecified fields are filled from predefined nominal defaults rather than through unconstrained generation.

Parsing is implemented as a bounded repair loop. If a candidate specification violates the schema or fails the analytic verifier, its diagnostic feedback is returned to the same interpretation operator:
\begin{equation}
S^{(k+1)}
=
\Lambda
\left(
p,\,
S^{(k)},\,
\mathcal{E}(S^{(k)})
\right),
\label{eq:parse}
\end{equation}
where $\mathcal{E}(S^{(k)})$ contains machine-readable violations together with actionable repair guidance. The LLM therefore performs semantic interpretation and repair, while the admissible specification space remains controlled by explicit constraints. Appendix~\ref{app:prompts} gives the parser, planner, and reflection prompts in condensed form.

\paragraph{Analytic Feasibility Verification}
\label{sec:verify}

The verifier determines whether a candidate can realize the communication condition it claims to test before simulator resources are committed. For a temporal fault $F_j$, realizability is evaluated against the freshness budget $B_a$ of the information stream consumed by application $a$. A fault is potentially consequential when
\begin{equation}
\overline{\Delta}(F_j)>B_a,
\end{equation}
where $\overline{\Delta}(F_j)$ is a lower bound on the peak Age of Information (AoI)~\cite{kaul2012realtime} induced by the fault. For a stream with broadcast period $T_s$ and nominal latency $L$, dropping $k$ consecutive messages gives
\begin{equation}
\overline{\Delta}=kT_s+L,
\end{equation}
whereas an additional communication delay $d$ gives
\begin{equation}
\overline{\Delta}=L+d.
\end{equation}
The true peak can exceed these bounds by up to one broadcast period, since that much time may already have passed since the last delivery; a fault that satisfies $\overline{\Delta}(F_j)>B_a$ is therefore guaranteed to breach the budget. If a requested drop is too weak, for example, the verifier can return the smallest count whose bound reaches the budget
\begin{equation}
k^\ast
=
\left\lceil
\frac{B_a-L}{T_s}
\right\rceil .
\label{eq:kstar}
\end{equation}

Semantic-content faults require a different test because they can remain apparently fresh. For a frozen SPaT value, incorrect MAP binding, or perturbed BSM position, the verifier checks whether the targeted information item is actually consumed by application $a$, whether the activation window overlaps a relevant decision period, and whether the requested corruption can produce $\beta(t)\neq W(t)$.

Additional checks cover fault-window validity, conflict timing, V2X equipage consistency, nominal communication settings, and roadway satisfiability. Blocking violations are returned through $\mathcal{E}$ to the parser. The same verification gate is later reused to eliminate infeasible mutants during Scenario Refinement. Appendix~\ref{app:checks} lists all check families.

\subsection{Module II: Retrieval-Augmented Scenario Grounding}
\label{sec:graph}

A verified specification describes what interaction should occur, but it is still symbolic. It does not yet identify where that interaction can physically occur, which concrete lanes correspond to semantic roles such as ``crossing from the left,'' or how the specified agents should behave. The second module therefore binds the specification to retrieved roadway, V2X, and behavioral knowledge~\cite{wu2026knowledge} and produces a replayable scenario program.

\paragraph{Signal-Semantic Road Retrieval}

A signal-semantic road graph $\mathcal{G}$ is constructed offline from the OpenDRIVE map of each simulator town. Unlike a conventional road graph that records only coarse topology or whether an intersection is signalized, $\mathcal{G}$ preserves controller identity, traffic-signal identity, junction connections, controlled movements, and signal groups. These quantities are required not only for geometric grounding but also because SPaT and MAP faults directly operate on them.

The basic retrieval unit is
\[
\ell
=
(\mathrm{junction},
 \mathrm{movement},
 \mathrm{signal\ group},
 \mathbf{f}_\ell),
\]
where $\mathbf{f}_\ell$ records geometric and traffic-control attributes. Given application $a$ and map query $Q$, the candidate set is
\begin{equation}
\mathcal{L}(S)
=
\left\{
\ell\in\mathcal{G}:
\mathrm{pre}(a)(\mathbf{f}_\ell)
\wedge
q_Q(\ell)
\right\},
\label{eq:locus}
\end{equation}
where $\mathrm{pre}(a)$ encodes the geometric preconditions of the target application and $q_Q$ applies the map requirements inferred from the prompt. A seeded deterministic rule selects $\ell^\ast$ from the feasible set. If no candidate exists, the failed constraint is returned to Module~I for repair.

\paragraph{V2X and Driver-Behavior Retrieval}

A second knowledge base contains the V2X semantics required to convert the traffic scene into a safety test. It represents the three message streams implemented by \sys{}: SPaT messages describing signal phase and timing, MAP messages describing intersection geometry and lane-to-signal-group associations, and BSMs describing the states of equipped vehicles.

An application registry maps each application $a$ to the information streams it consumes, its geometric preconditions, and its information-quality budgets. A fault-pattern library maps natural-language communication degradations to the protocol-level fault operators implemented by the runtime.

Driver behavior is instantiated through parameterized templates. For vehicle $i$,
\begin{equation}
\theta_i
=
w_E
\odot
T_{\mathrm{beh}}(c_i,b_i),
\label{eq:behavior}
\end{equation}
where $T_{\mathrm{beh}}$ maps vehicle class $c_i$ and behavioral character $b_i$ to parameters including desired speed, time headway, standstill clearance, reaction time, comfortable deceleration, and signal compliance. Environmental weighting $w_E$ adjusts these parameters according to $E$.

\paragraph{Lane-Level Scene Planning}
\label{sec:planner}

The retrieved movement unit determines a feasible junction, but each semantic agent role must still be mapped onto an actual lane. The executor queries the selected junction in the live simulator and constructs an admissible approach set $\mathcal{J}(\ell^\ast)$ containing lane identifiers, movement directions, and measured headings relative to the ego approach.

A constrained planning operator assigns agents to these retrieved approaches:
\begin{equation}
\mathcal{Z}
=
\Lambda_{\mathrm{plan}}
\left(
S,\mathcal{J}(\ell^\ast)
\right).
\label{eq:plan}
\end{equation}
The LLM may therefore resolve semantic relationships such as ``oncoming'' or ``crossing from the left,'' but it cannot invent lanes or roadway geometry.

Each assignment is then validated deterministically. Using the specified spawn distance and initial speed, the validator checks whether the relevant vehicles can reach the conflict region within the required encounter window. Invalid assignments are rejected and deterministic grounding heuristics are used as a fallback where necessary.

\paragraph{Scenario Program Assembly}

Finally, the verified specification, retrieved movement, lane assignments $\mathcal{Z}$, and behavior parameters are combined through
\begin{equation}
P
=
\mathrm{bind}
\left(
S,\ell^\ast,\mathcal{Z},\{\theta_i\}
\right).
\label{eq:program}
\end{equation}
The resulting program contains the agent routes, signal plan $\Psi$, V2X message streams, communication settings, event-anchored fault schedule, and monitoring configuration required for execution.

Importantly, $P$ is typed scenario data consumed by a fixed runtime engine; it is not LLM-generated simulator source code. This makes each test reproducible, allows later refinement operators to manipulate the specification symbolically, and preserves traceability between the executable scenario and the original prompt or retrieved knowledge.

\subsection{Module III: Automated V2X Scenario Synthesis}
\label{sec:grounding}

The third module executes the grounded scenario program as a coupled physical--information process. Unlike a conventional traffic simulation that models vehicle motion alone, \sys{} synchronously advances the traffic state, signal state, V2X messages, receiver-side estimated state, application decisions, and paired ego outcomes. Execution produces the synchronized trace
\begin{equation}
\mathcal{T}
=
\mathrm{Exec}(P).
\end{equation}

\paragraph{Synchronous Physical--Information Execution}
\label{sec:execution}

Simulation proceeds synchronously with fixed step $\Delta t=50$\,ms. At each tick, messages are sampled from the current world, propagated through the communication channel, degraded by any active faults, and incorporated into the estimated information state. The application then acts on that estimate:
\begin{equation}
\beta_{t+\Delta t}
=
\mathcal{B}
\left(
\beta_t,\,
(\Phi_F\circ\Gamma\circ\Sigma)(W_t)
\right),
\end{equation}
followed by
\begin{equation}
W_{t+\Delta t}
=
\Phi_{\mathrm{sim}}
\left(
W_t,\,
\pi_a(\beta_t,x_t^{\mathrm{msg}}),\,
u_t,\,
\Psi_t
\right),
\label{eq:step}
\end{equation}
where $u_t$ contains the controls of the non-ego agents.

Vehicle routes are instantiated on the live simulator map, initial conditions are stabilized before the critical encounter, conflict-relevant agents follow the scripted controllers parameterized by $\theta_i$, and background traffic is controlled through a seeded Traffic Manager. The signal plan is aligned with the intended encounter so that event-anchored communication faults occur relative to the corresponding traffic or signal event rather than at an arbitrary simulation time.

\paragraph{Spatial V2X Communication Channel}
\label{sec:channel}

The communication operator $\Gamma$ models delivery as a function of the instantaneous physical relationship between transmitter and receiver. For each link, latency $L$ and delivery probability $p_{\mathrm{del}}$ are obtained from received signal strength:
\begin{equation}
(L,p_{\mathrm{del}})
=
g(\mathrm{RSSI}),
\qquad
\mathrm{RSSI}
=
P_{\mathrm{tx}}
-
10n\log_{10}d
-
B_{\mathrm{blk}}(x_{\mathrm{tx}},x_{\mathrm{rx}}),
\label{eq:rssi}
\end{equation}
where $P_{\mathrm{tx}}$ is transmit power, $n$ is the path-loss exponent, $d$ is transmitter--receiver distance, and $B_{\mathrm{blk}}$ is the height-aware building-blockage term. The mapping $g$ increases latency and reduces delivery probability as signal strength deteriorates.

Roadside units are placed at physical signal-pole locations, so communication quality evolves with vehicle motion, urban geometry, and transmitter placement rather than being represented by one scenario-wide packet-loss rate. Appendix~\ref{app:qual} gives the complete model and its default parameters.

\paragraph{Semantic V2X Fault Injection}
\label{sec:faults}

After nominal channel propagation, the fault engine $\Phi_F$ applies protocol-level message faults. Seven operators are implemented across three failure families. Delivery faults include \texttt{drop}, \texttt{burst\_drop}, and \texttt{delay}; the timestamp fault \texttt{skew} changes the reported generation time; and semantic-content faults include \texttt{freeze}, \texttt{rebind}, and \texttt{pos\_perturb}. Appendix~\ref{app:faults} defines each operator and its parameters.

These operators alter different parts of the information chain. Delivery and timing faults primarily change message availability or apparent age, whereas semantic faults may preserve normal delivery and fresh timestamps while changing the represented state. Fault windows are anchored to traffic or signal events and converted into deterministic simulation-time intervals, allowing identical tests to be reproduced across paired executions and subsequent replays.

\paragraph{Paired Counterfactual Execution}

A shadow ground-truth ego is propagated alongside the message-driven ego. Both use the same application, controller, route, initial condition, traffic scene, and simulation seed. Their only systematic difference is the information provided to the application: the message-driven ego acts on $\beta(t)$, whereas the ground-truth ego acts on the corresponding state derived directly from $W(t)$.

The shadow ego has no physical presence and therefore cannot alter surrounding traffic. This paired construction provides the counterfactual required by Eq.~(\ref{eq:admission}). Each simulation tick records physical state, signal state, message transmissions and receptions, reconstructed information, application decisions, and both ego outcomes. The resulting trace $\mathcal{T}$ therefore provides an auditable record of the complete chain from communication degradation to behavioral consequence.

\subsection{Module IV: Scenario Refinement}
\label{sec:refine}

\begin{wrapfigure}{r}{0.5\textwidth}
\vspace{-22pt}
\begin{minipage}{\linewidth}
\begin{algorithm}[H]
\caption{Scenario refinement by search and reflection.}
\label{alg:refine}
\footnotesize
\begin{algorithmic}[1]
\Require prompt $p^{(0)}$, goal $g$, budgets $N_r$, $N_s$, elite size $K$
\Ensure minimal witnesses, or diagnoses $\{\mathcal{D}^{(r)}\}$
\For{$r=0,\dots,N_r-1$} \Comment{reflection}
    \State $S \Gets \mathcal{C}(p^{(r)})$;\; $\mathcal{T}\Gets\mathrm{Exec}(P)$ \Comment{Mod.\ I--III}
    \IfThen{$A(S)=1$}{\Return $\textsc{DeltaDebug}(S)$}
    \State $\mathcal{S} \Gets \{S\}$;\; $\mathcal{H} \Gets \emptyset$ \Comment{elite, admitted}
    \For{$b=1,\dots,N_s$} \Comment{search}
        \State $S' \Gets \textsc{Mutate}\big(\textsc{Sample}_w(\mathcal{S})\big)$
        \IfThen{$\neg\textsc{Verify}(S')$}{\textbf{continue}} \Comment{gated}
        \State execute $S'$; compute $m(S')$
        \IfThen{$A(S')=1$}{$\mathcal{H} \Gets \mathcal{H}\cup\{S'\}$} \Comment{catalog}
        \State $\mathcal{S} \Gets \textsc{Top}_K(\mathcal{S}\cup\{S'\})$ by $w$
    \EndFor
    \IfThen{$\mathcal{H}\neq\emptyset$}{\Return $\textsc{DeltaDebug}(\mathcal{H})$}
    \State $\mathcal{D}^{(r)} \Gets \textsc{Digest}(\mathcal{T},\ldots)$
    \State $p^{(r+1)} \Gets \Lambda_{\mathrm{ref}}(p^{(r)},\mathcal{D}^{(r)},g)$ \Comment{Eq.~(\ref{eq:reflect})}
\EndFor
\State \Return $\{\mathcal{D}^{(r)}\}_{r=0}^{N_r-1}$ \Comment{no admission}
\end{algorithmic}
\end{algorithm}
\end{minipage}
\vspace{-6pt}
\end{wrapfigure}
A single execution may expose an admitted failure, but many initial tests are too weak, mistimed, or otherwise poorly configured to make the intended communication mechanism consequential. Scenario Refinement therefore closes the loop between simulation evidence and scenario generation.

Two complementary refinement processes are used. An inner \emph{specification-space search} varies bounded properties of the current grounded scenario while preserving its semantic intent. An outer \emph{prompt-space reflection} revises the natural-language test description when the execution evidence indicates that the original intent itself prevents the desired mechanism from becoming consequential. Algorithm~\ref{alg:refine} summarizes both loops.

\paragraph{Execution Diagnosis and Causal Admission}
\label{sec:divergence}

The refinement monitor first identifies intervals in which a safety-relevant estimated information item differs from its corresponding ground-truth state,
\begin{equation}
\beta_i(t)\neq W_i(t).
\end{equation}
Each divergence episode is related to the next application decision through a \emph{hazard slack} $\delta$. Negative slack indicates that a decision was made while the estimated state was incorrect, whereas a small positive value denotes a near miss. Expected absences, such as missing BSMs from unequipped vehicles, are labeled separately as structural silent-by-absence conditions.

The paired rollout then evaluates the admission criterion of Eq.~(\ref{eq:admission}). If both egos violate the safety criterion, the scenario is treated as non-attributable because the underlying traffic interaction itself is unsafe. If neither fails, the trace remains useful for determining how close the communication-induced divergence came to changing the outcome. Only
\[
V_{\mathrm{msg}}=1,
\qquad
V_{\mathrm{gt}}=0
\]
constitutes a communication-caused failure.

\paragraph{Specification-Space Search}
\label{sec:coverage}

When the current specification is valid but does not produce an admitted failure, the inner loop searches a bounded neighborhood of that scenario. Candidate mutations modify parameters including fault type, magnitude and activation window, vehicle distance and speed, and V2X penetration. These changes preserve the high-level test intent while altering how strongly the information degradation overlaps the safety-critical interaction.

Every proposed mutant is first passed through the analytic verifier. Feasible mutants are executed and ranked using failure margin $m(S)$. Candidates closer to failure receive greater probability of producing descendants according to
\begin{equation}
w(S)
=
\frac{1}{1/2+m(S)}.
\label{eq:search-weight}
\end{equation}
Thus, execution-derived near misses focus subsequent search without replacing the causal admission test itself.

Search is also coverage-oriented. Admitted scenarios are cataloged by mechanism dimensions including application, fault family, V2X penetration, and conflict geometry, discouraging repeated discovery of equivalent failures.

Each admitted scenario is then minimized by delta debugging, a greedy descent: at every step it tries each single weakening of the fault vector (removing a fault, halving its magnitude, or shortening its window) and keeps the first that preserves admission under the same paired causal test, stopping when none does. The result approximates the minimal-witness objective of Eq.~(\ref{eq:minimal-witness}): it is 1-minimal, in that no single fault can be removed or weakened without losing the failure, and it yields a replayable scenario containing only the communication degradation necessary for the failure.

\paragraph{Prompt-Space Reflection}

Bounded specification mutation cannot repair every unsuccessful test. In some cases, the natural-language intent itself makes the requested failure mechanism irrelevant. For example, a prompt may activate a SPaT freeze only after the ego has already completed the safety-critical decision. Increasing the magnitude of that fault cannot make it causal without changing its temporal relationship to the described event.

The outer loop therefore constructs an evidence digest $\mathcal{D}(\mathcal{T}^{(r)})$ containing the paired outcomes, divergence episodes, hazard slack, decision-epoch AoI, failure margins, and verifier diagnostics. Following verbal self-reflection and self-refinement for LLM agents~\cite{shinn2023reflexion,madaan2023selfrefine}, a dedicated reflection operator then updates the prompt:
\begin{equation}
p^{(r+1)}
=
\Lambda_{\mathrm{ref}}
\left(
p^{(r)},
\mathcal{D}(\mathcal{T}^{(r)}),
g
\right).
\label{eq:reflect}
\end{equation}

The purpose of reflection is to diagnose why the intended mechanism failed to affect the application and revise the test description accordingly. It may, for example, move a fault window so that it overlaps the decision, extend a semantic corruption through a signal transition, or modify a described geometry that prevents communication exposure.

The reflected prompt receives no privileged execution path. It re-enters Module~I and is parsed and analytically verified, is grounded again through Module~II, and is executed through Module~III under the same causal admission criterion. Reflection therefore cannot directly manipulate simulator state or force a violation; it can only propose a revised test whose validity and causal effect must again be established by the complete pipeline.

\section{Experimental Results}
\label{sec:experiments}

The experiments evaluate the three main capabilities of \sys{}: language-driven V2X scenario generation, causal verification of communication-induced failures, and adaptive refinement of ineffective tests. A 100-scene benchmark first evaluates whether natural-language scenarios can be generated reliably without producing false failures. Communication-fault experiments then test whether the paired execution can distinguish communication-induced failures from unsafe traffic interactions. Finally, a 50-scene refinement study evaluates whether parameter search and LLM self-reflection can systematically turn ineffective tests into causally verified failures. Ablation studies isolate the contribution of the key components.

\subsection{Experiment Setup}
\paragraph{Implementation and Protocol}
All experiments use the complete \sys{} pipeline with synchronous CARLA simulation across eight towns, except the specification-space search runs of the refinement study, which use a lightweight stub backend. GPT-4o~\cite{gpt4o} is used for language parsing, planning, and reflection with temperature 0.2 and a fixed sampling seed. Scenario seeds control vehicle equipage, communication loss, RSU selection, vehicle blueprints, and retrieval tie-breaking, enabling deterministic replay. For each scene, \sys{} records the generated specification, grounding results, execution traces, safety outcomes, and communication states. The evaluation includes 112 live CARLA scenes; 110 prompts compile on the first attempt and two require verifier-guided correction. Unless otherwise stated, results are reported from deterministic single-seed runs. Appendix~\ref{app:repro} summarizes the configuration of each experiment.

\paragraph{Safety Applications}
We evaluate six V2X safety applications, summarized in Table~\ref{tab:budgets}. RLVW and GLOSA are \emph{signal applications} that primarily use infrastructure-to-vehicle information, while IMA, LTA, FCW, and EEBL are \emph{conflict applications} that primarily use vehicle-to-vehicle BSMs. This distinction captures two important V2X settings: infrastructure information can be shared by multiple vehicles, whereas a missing BSM can make a conflicting vehicle invisible to the receiving application. Appendix~\ref{app:apps} gives each application's decision rule and violation criterion.

\begin{table*}[t]
\centering\footnotesize
\caption{V2X safety applications and their message requirements (SAE J2945/1~\cite{sae-j2945-1_2020}). Age budgets bound the staleness of the consumed SPaT or BSM stream; ``--'' marks a quantity the application does not use.}
\label{tab:budgets}
\begingroup
\setlength{\tabcolsep}{5pt}
\setlength{\aboverulesep}{0pt}
\setlength{\belowrulesep}{0pt}
\renewcommand{\arraystretch}{1.22}
\begin{tabularx}{\textwidth}{ll >{\raggedright\arraybackslash}X ccc}
\toprule
& & & \multicolumn{2}{c}{\textbf{Age budget (s)}} & \textbf{Position} \\
\cmidrule(lr){4-5}
\textbf{Application} & \textbf{Full name} & \textbf{Function} & \textbf{SPaT} & \textbf{BSM} & \textbf{error (m)} \\
\midrule
\rowcolor{TabGroup}\multicolumn{6}{l}{\textbf{Signal applications}\enspace\textit{infrastructure-to-vehicle (SPaT/MAP)}} \\
\hspace{0.7em}\textbf{RLVW} & Red-Light Violation Warning & \textcolor{black!65}{Warns against entering on red} & 1.0 & \textcolor{black!35}{--} & \textcolor{black!35}{--} \\
\hspace{0.7em}\textbf{GLOSA} & Green Light Optimal Speed Advisory & \textcolor{black!65}{Recommends speed for a green arrival} & 1.0 & \textcolor{black!35}{--} & \textcolor{black!35}{--} \\
\rowcolor{TabGroup}\multicolumn{6}{l}{\textbf{Conflict applications}\enspace\textit{vehicle-to-vehicle (BSM)}} \\
\hspace{0.7em}\textbf{IMA} & Intersection Movement Assist & \textcolor{black!65}{Warns of crossing-vehicle conflicts} & \textcolor{black!35}{--} & 0.5 & 1.5 \\
\hspace{0.7em}\textbf{LTA} & Left Turn Assist & \textcolor{black!65}{Warns of conflicting opposing traffic} & 1.0 & 0.5 & 1.5 \\
\hspace{0.7em}\textbf{FCW} & Forward Collision Warning & \textcolor{black!65}{Warns of a slower or stopped lead vehicle} & \textcolor{black!35}{--} & 0.3 & 1.5 \\
\hspace{0.7em}\textbf{EEBL} & Emergency Electronic Brake Light & \textcolor{black!65}{Relays a hard-braking event ahead} & \textcolor{black!35}{--} & 0.3 & \textcolor{black!35}{--} \\
\bottomrule
\end{tabularx}
\endgroup
\end{table*}

\paragraph{Experiment Groups}

\begin{figure*}[!ht] \centering 
\includegraphics[width=0.98\textwidth]{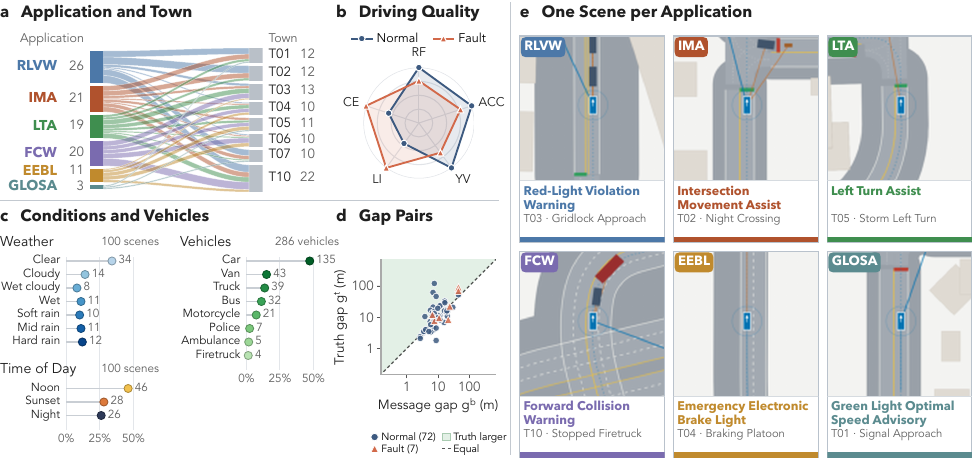} \caption{100 generated scenes. \textbf{a}, Application and town. \textbf{b}, Driving quality (RF, route following; ACC, acceleration; YV, yaw rate; LI, lane invasion; CE, communication exposure). \textbf{c}, Scene conditions. \textbf{d}, Minimum gap. \textbf{e}, Examples.} \label{fig:track-gen}
\end{figure*} 

The evaluation contains three main experiment groups. First, a 100-scene normal-traffic benchmark tests whether natural-language descriptions can be converted into valid V2X scenarios without producing false communication-induced failures. The prompts describe only the traffic scene and environment; the exercised application, town, and nominal communication configuration are inferred or assigned by the pipeline. The benchmark spans all six applications and all eight CARLA towns, together with varied weather, times of day, and vehicle classes (Fig.~\ref{fig:track-gen}a,c). Second, a communication-fault suite evaluates whether \sys{} can distinguish failures caused by degraded V2X information from those caused by the underlying traffic interaction. Third, the suite is extended to 50 scenarios to evaluate adaptive refinement through specification-space search and LLM self-reflection. Corridor, multi-CV, and channel-sensitivity experiments are treated as supplementary studies.

\paragraph{Evaluation Metrics}
Metrics are organized at four levels. The \emph{run level} determines whether communication caused a safety failure. Each executed scenario is classified as \textsc{Admitted}, \textsc{Non-attributable}, or \textsc{Clean}. \textsc{Admitted} means the degraded-information rollout violates the safety criterion while the matched non-degraded rollout does not. \textsc{Non-attributable} means both rollouts violate, indicating that the failure cannot be attributed to communication. \textsc{Clean} means neither rollout violates. Scenarios rejected before execution are labeled \textsc{Gated}.

For scenarios with a measurable conflict gap, we also report
\begin{equation}
\Delta g = g^{\mathrm{t}} - g^{\mathrm{b}},
\label{eq:effect}
\end{equation}
where $g^{\mathrm{t}}$ and $g^{\mathrm{b}}$ are the minimum conflict gaps under non-degraded and degraded V2X information, respectively. Positive $\Delta g$ indicates a loss of physical separation caused by degraded information.

The \emph{estimated-state level} measures how strongly the vehicle's perceived state deviates from the physical state. We report hazard slack, decision-epoch Age of Information (AoI), and the type and duration of each estimate--truth divergence. The \emph{search level} measures refinement efficiency. We report the number of admitted failures, attribution rate, feasibility-gate rejection rate, executions required to reach the first admission, and minimal-witness size after delta debugging. The \emph{realization level} verifies that the generated scene was executed as intended. Parse success, spawn audits, and route completion are used to detect realization artifacts, while the 100-scene normal-traffic benchmark serves as a negative control for false admissions.

\subsection{Scenario Generation at Scale}
\label{sec:lang-results}
\label{sec:diverse}

\begin{figure*}[!t] \centering 
\includegraphics[width=0.98\textwidth]{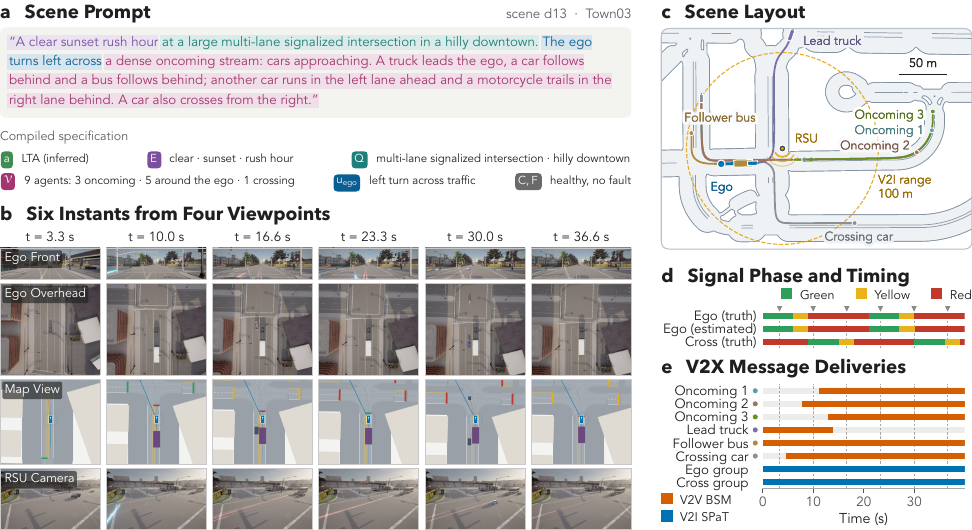} 
\caption{A nine-agent left turn in Town03 under a healthy channel. \textbf{a}, Prompt and compiled specification. \textbf{b}, Six instants from four viewpoints. \textbf{c}, Agent paths and V2I range. \textbf{d}, True and estimated signal phases. \textbf{e}, Message deliveries.} 
\label{fig:scene-d13} 
\end{figure*}

The first experiment evaluates whether plain traffic language can drive the full pipeline at scale. Figure~\ref{fig:track-gen} summarizes the 100-scene normal-traffic benchmark. The prompts describe only the traffic scene and environment: they do not name the exercised V2X application, the CARLA town, or any communication fault. The application is inferred by the parser, feasible towns are identified through retrieval, and the benchmark harness assigns the least-used feasible town. As a result, coverage is measured rather than hand-designed. The benchmark covers six V2X applications across eight CARLA towns, with various weather conditions, times of day, vehicle classes, and traffic densities ranging from a single ego vehicle to eleven agents.

Generation is both reliable and faithful. All 100 scenes compile into valid and feasible specifications within at most two parse attempts, with an average of 1.58 attempts per scene, and every scene is grounded to a feasible movement unit and executed to completion. The only recurring compile issue is an out-of-vocabulary \code{maneuver} field, which is corrected on retry. Realization quality is also high: 365 of 367 vehicle spawns are clean, all 2048 checked lane markings match the authored OpenDRIVE map, and 253 of 254 vehicles remain on route.

Most importantly, the normal-traffic corpus yields zero admitted communication-induced failures under healthy communication, confirming that \sys{} does not manufacture failures. The generated scenes also remain physically well behaved, with zero collisions, zero out-of-road distance, and sub-0.15\,m tracking error in both rollouts. The paired executions further provide a scene-level measure of information impact. Among the 79 scenes with finite conflict gaps in both rollouts, the ground-truth ego maintains more than 0.5\,m additional separation in 33 scenes, with a median gain of 8.5\,m and a maximum of 112.3\,m. Figure~\ref{fig:scene-d13} shows one representative dense control scene in detail. Overall, these results show that plain traffic language is sufficient to drive \sys{} to broad and faithful V2X scenario generation without introducing spurious failures.

\subsection{Corridor Traversal along an RSU Chain}
\label{sec:corridor}

\begin{figure*}[!t]
  \centering
  \includegraphics[width=0.98\textwidth]{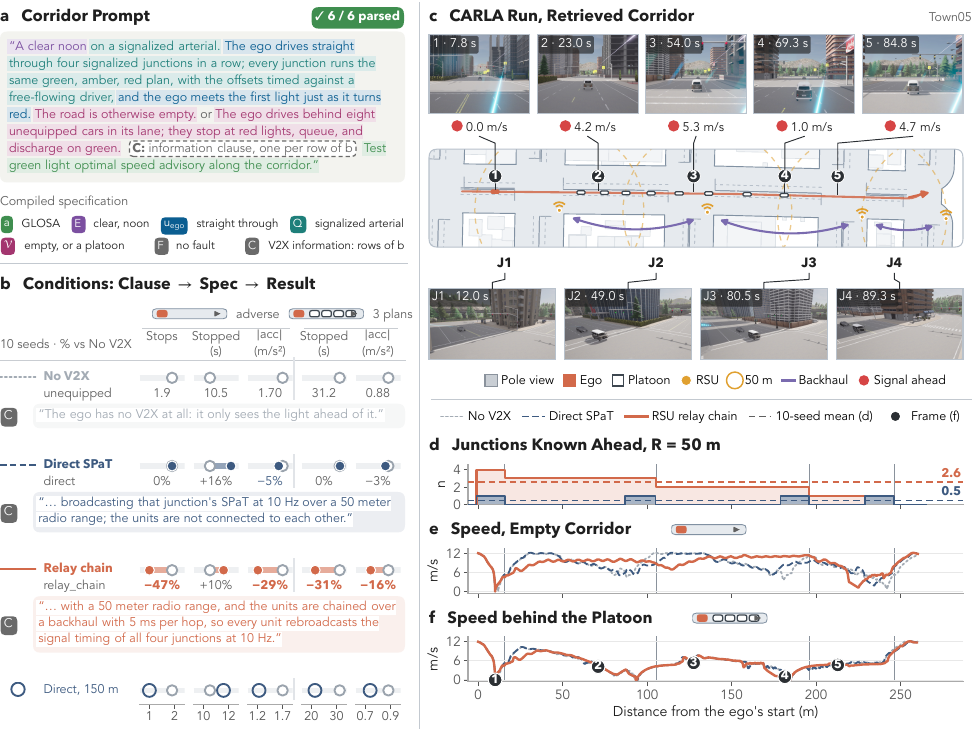}
\caption{Corridor traversal along an RSU chain. \textbf{a}, Shared prompt; $C$ is the condition-specific clause. \textbf{b}, Clause, parsed value, and 10-seed results per condition, relative to No V2X. \textbf{c}, Camera frames on the retrieved corridor. \textbf{d}, Junctions with known SPaT. \textbf{e},\textbf{f}, Ego speed without and with traffic.}
  \label{fig:track-corridor}
\end{figure*}

The corridor study measures what infrastructure ahead of the trip buys: a connected vehicle's information horizon need not end at its own radio. Like the main benchmark, it is driven from language (Fig.~\ref{fig:track-corridor}a). One prompt describes the corridor, four signalized junctions with a shared 6/3/12\,s plan and offsets timed against a free-flowing driver, and differs between runs only in whether the road is empty or the ego follows eight unequipped cars, and in one information clause (Fig.~\ref{fig:track-corridor}b): no V2X; \emph{direct} SPaT, where each junction's RSU broadcasts only its own timing over a 50\,m radio; and an \emph{RSU relay chain}, where backhaul-connected RSUs rebroadcast the whole corridor's SPaT at 5\,ms per hop, so one reachable RSU delivers every junction ahead. All six prompts compile to the intended condition on the first parse; the verifier's only repairs lengthen a too-short episode. Retrieval binds the corridor to a Town05 arterial with junctions 91, 88 and 54\,m apart, and the harness then sweeps the offset plan (aligned, green wave, adverse), the radio range (50, 100, 150\,m) and 10 seeds: 420 CARLA runs with no red-light entry and no collision.

The chain's value concentrates where the radio is weakest. At $R = 50$\,m, below the junction spacing, direct coverage is intermittent, and the relay lifts the mean number of junctions known ahead from 0.5 to 2.6 (Fig.~\ref{fig:track-corridor}d). With the horizon comes smoothness. Against adverse offsets on the empty corridor, GLOSA over the chained SPaT cuts stops from 1.9 to 1.0 and the mean acceleration magnitude from 1.70 to 1.21\,m/s$^2$ relative to no V2X, while direct SPaT at the same range changes little (1.9 stops, 1.62\,m/s$^2$; Fig.~\ref{fig:track-corridor}b); between J1 and J3 the relayed ego holds about 10\,m/s where the others brake and re-accelerate (Fig.~\ref{fig:track-corridor}e). Travel time stays signal-bound (47.4\,s without V2X versus 46.6\,s with the relay chain), and the one stop that remains, at the first light, which every condition meets on red, is slightly longer than the no-V2X ego's total stopped time (11.6 versus 10.5\,s): the corridor's benefit is fewer stops and gentler control, not higher speed. The chain at $R = 50$\,m matches direct SPaT at three times the range (1.0 stops, 1.23\,m/s$^2$ at 150\,m); infrastructure cooperation substitutes for radio range.

With the eight-car platoon ahead, the queue sets the pace and travel time is the same in all three conditions (98\,s, averaged over the offset plans). What changes is how that time is spent: the relay chain cuts the ego's time stopped from 31.2 to 21.5\,s and its mean acceleration magnitude from 0.88 to 0.74\,m/s$^2$, while direct SPaT, whose short radio rarely sees past the queue, changes neither (31.3\,s, 0.86\,m/s$^2$; Fig.~\ref{fig:track-corridor}b,f). The effect is largest under the green-wave plan, where idling falls from 34.0 to 13.3\,s: the ego glides toward queues it knows will discharge instead of braking into them. In traffic, corridor information buys smoothness rather than time.

\subsection{Cooperative Scenarios with Multiple Connected Vehicles}
\label{sec:multicv}

\begin{figure*}[!t]
  \centering
  \includegraphics[width=0.98\textwidth]{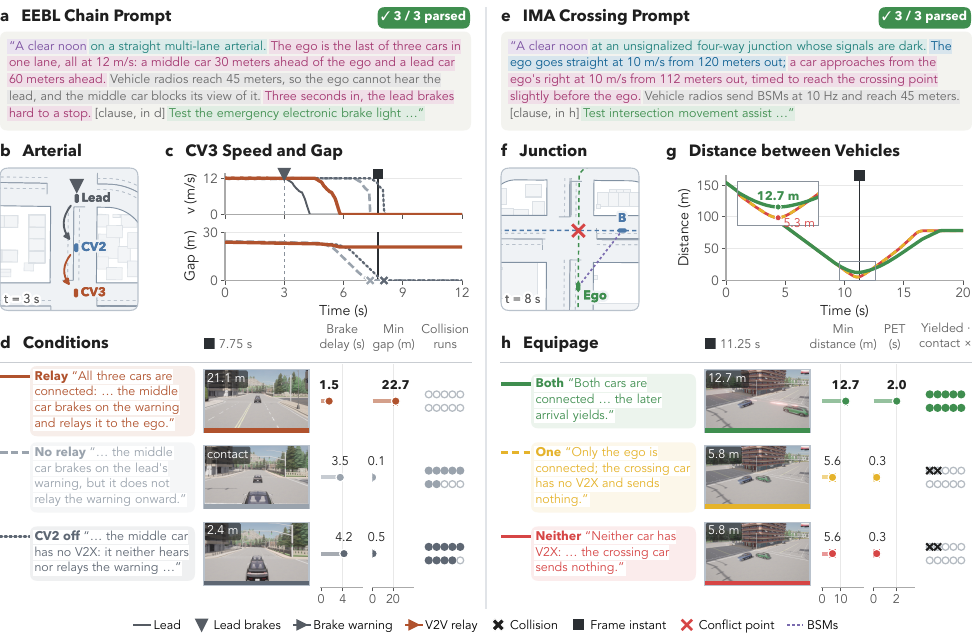}
\caption{Multi-CV cooperation: EEBL relay chain (a--d) and IMA mutual yield (e--h). \textbf{a},\textbf{e}, Prompts. \textbf{b},\textbf{f}, Retrieved scenes. \textbf{c}, CV3 speed and gap. \textbf{g}, Distance between the crossing vehicles. \textbf{d},\textbf{h}, Clause, CARLA frame, and 10-seed results per condition.}
  \label{fig:track-multicv}
\end{figure*}

The multi-CV study makes cooperation itself the object under test: with every connected vehicle owning an estimated-state store and a decision policy, what two vehicles achieve together becomes measurable (Fig.~\ref{fig:track-multicv}). Both experiments are again stated in language, one prompt per condition, and all six prompts compile to the intended condition on the first parse; retrieval binds the brake-warning chain to a straight Town05 arterial and the crossing to Town05 junction 1722, and each condition runs for 10 seeds. In the EEBL V2V relay chain, a lead vehicle brakes hard beyond the third vehicle's radio range and behind an occluding second vehicle, so whether CV3 learns of the event in time depends on whether CV2 rebroadcasts the warning vehicle-to-vehicle. With the V2V relay, CV3 brakes 1.49\,s after the lead and keeps a 22.7\,m minimum gap, with no collision in 10 seeds. Without it, CV3 waits 3.52\,s for visual evidence, the gap closes to 0.1\,m and 7 of 10 runs end in a rear-end collision; with CV2 unequipped, CV3 brakes after 4.21\,s and 9 of 10 runs collide (Fig.~\ref{fig:track-multicv}c,d). The V2V relay hop converts an occluding vehicle into infrastructure worth 2.0\,s of warning.

The IMA mutual-yield experiment measures the other face of cooperation. Two crossing vehicles timed to conflict interleave cleanly when both are equipped: a 12.7\,m closest approach and a 2.04\,s post-encroachment time, with the later arrival yielding in 10 of 10 seeds. With one or neither equipped, the encounter degenerates to a 5.6\,m near miss with a 0.28\,s post-encroachment time and contact in 2 of 10 seeds, identically in both cases (Fig.~\ref{fig:track-multicv}g,h). Cooperation is pairwise: one-sided equipage buys nothing when the counterpart is silent. This restates the corpus's silent-miss finding at the scale of two deciding vehicles.

\subsection{Scenario Refinement at Scale}
\label{sec:refine-results}

\begin{figure*}[!t]
\centering
\includegraphics[width=0.98\textwidth]{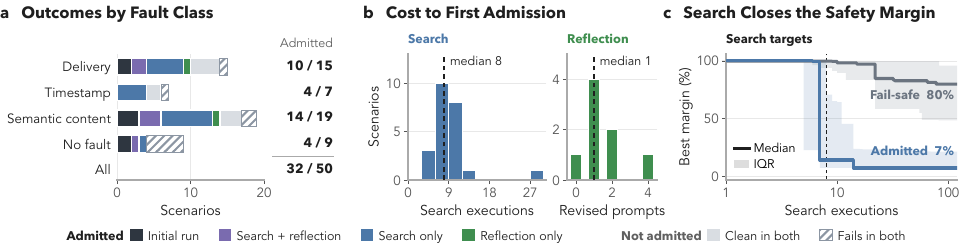}
\caption{Scenario refinement on the 50-scene fault suite. \textbf{a}, Outcome by fault class. \textbf{b}, Executions or prompt revisions to first admission. \textbf{c}, Median best margin during search (shaded: IQR).}
\label{fig:refinement}
\end{figure*}

The refinement experiment evaluates whether ineffective V2X tests can be systematically improved through specification-space search and LLM self-reflection. The fault suite is expanded from 12 to 50 prompts spanning six applications and diverse communication faults, including intentionally weak, mistimed, and ineffective cases. All scenarios are first executed without refinement. The 43 non-admitted cases are then refined using specification-space search with a budget of 120 candidate mutants per scene and LLM self-reflection with up to five revisions. To measure each mechanism's contribution, the two loops of Algorithm~\ref{alg:refine} are applied separately to every target: search mutates the compiled scenario, and reflection revises the prompt, with each revision compiled and executed once.

The initial execution admits only 7 of the 50 scenarios. Most missing, delayed, or stale messages do not produce failures because the application can detect or tolerate them, whereas incorrect but apparently valid information is more likely to alter vehicle behavior. This result reinforces an important point: injecting a communication fault does not by itself create a meaningful V2X safety test.

Across the 50 scenarios, refinement increases admitted failures from 7 to 32 (Fig.~\ref{fig:refinement}a). Specification-space search finds failures in 23 of the 43 refinement targets, with a median of 8 executions to the first admission and no successful case requiring more than 27 executions. LLM self-reflection rescues 8 targets, typically within one revised prompt (Fig.~\ref{fig:refinement}b). Six of these are also reachable through parameter search, while two require semantic changes beyond the bounded search space. The two refinement mechanisms are therefore complementary: search adjusts quantitative conditions such as fault timing and interaction severity, whereas self-reflection repairs semantically or structurally ineffective test intents.

For scenarios successfully converted by search, the best safety margin decreases to a median of 7\% of its initial value, compared with 80\% for scenarios that remain fail-safe (Fig.~\ref{fig:refinement}c). Delta debugging further reduces each admitted case to a 1-minimal witness with a single fault. Of the 18 scenarios that remain non-admitted, most either fail under both communication conditions and are therefore non-attributable, or remain fail-safe despite refinement. These results show that \sys{} can systematically improve ineffective tests without forcing every scenario into an unsafe outcome.

\begin{figure*}[!t]
\centering
\includegraphics[width=0.98\textwidth]{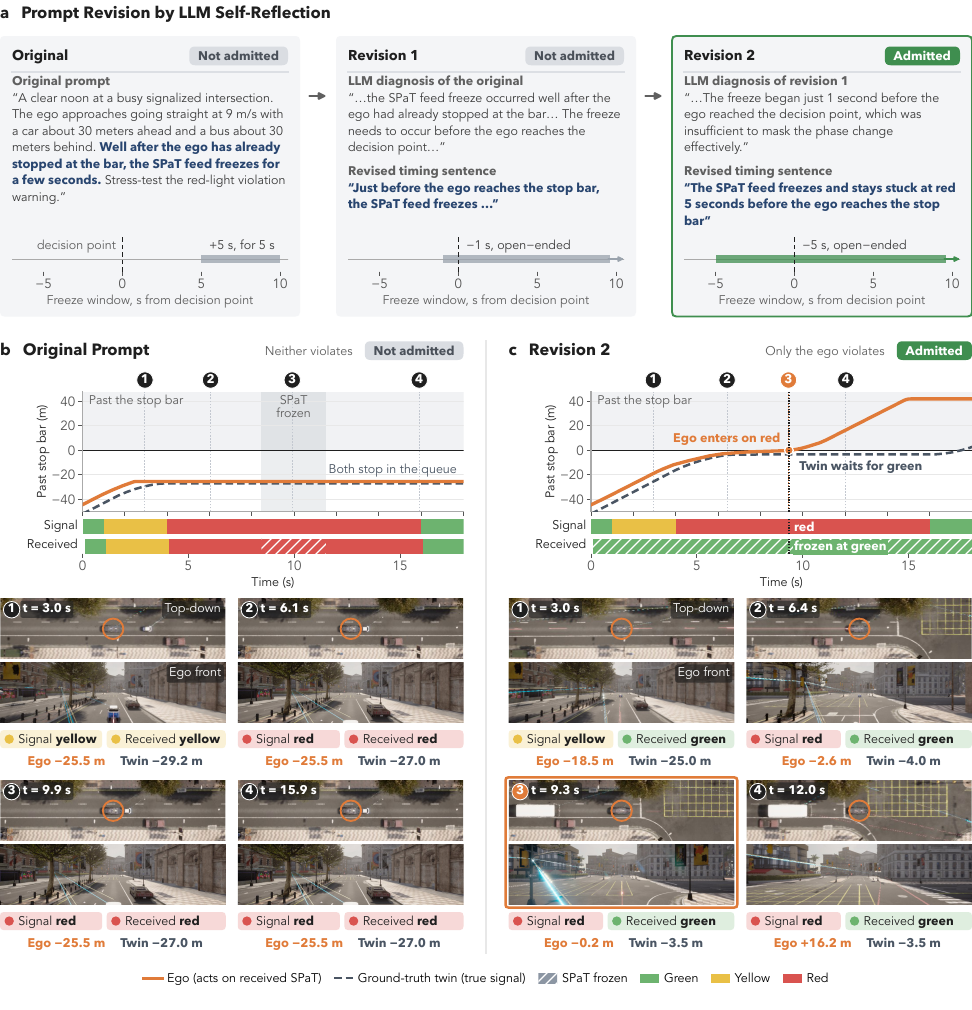}
\caption{Refinement example \code{rlvw\_late\_freeze}. \textbf{a}, Prompt revisions by LLM self-reflection. \textbf{b},\textbf{c}, Paired runs of the original and revised prompts: ego and ground-truth twin, true and received signal phases, and CARLA frames.}
\label{fig:refinement-case}
\end{figure*}

Figure~\ref{fig:refinement-case} illustrates how self-reflection repairs an ineffective test. In \code{rlvw\_late\_freeze}, the original SPaT freeze begins after the ego has already made its stopping decision, so the scenario remains clean. Using the execution evidence, the LLM identifies the mistimed fault and revises the scenario so that the frozen SPaT persists through the critical decision window. After refinement, the message-driven ego enters the intersection on red while the matched non-degraded ego stops, converting the original scenario into a causally verified communication-induced failure.

\subsection{Ablation and Sensitivity Studies}
\label{sec:ablation}
\label{sec:channel-study}

Table~\ref{tab:ablation} evaluates the contribution of each major pipeline component by removing it individually and comparing the resulting performance with the full system. The ablations test launch conditioning, height-aware blockage, the ground-truth rollout, the feasibility gate, the scene planner, and prompt reflection.

\begin{table}[!t]
\centering\footnotesize
\caption{Ablation of \sys{} components. Each block removes one component from the full pipeline; $\uparrow$/$\downarrow$ marks whether a higher or lower value is better, and the better value is in bold.}
\label{tab:ablation}
\begingroup
\setlength{\tabcolsep}{3.5pt}
\setlength{\aboverulesep}{0pt}
\setlength{\belowrulesep}{0pt}
\renewcommand{\arraystretch}{1.25}
\begin{tabularx}{\linewidth}{>{\raggedright\arraybackslash}X c >{\columncolor{TabFull}}r r @{\hspace{0.9em}{\color{black!25}\vrule width 0.4pt}\hspace{0.9em}} >{\raggedright\arraybackslash}X c >{\columncolor{TabFull}}r r}
\toprule
\textbf{Component and metric} & & \textbf{With} & \textbf{Without} & \textbf{Component and metric} & & \textbf{With} & \textbf{Without} \\
\midrule
\multicolumn{4}{l}{\cellcolor{TabGroup}\textbf{Ground-truth rollout}} & \multicolumn{4}{l}{\cellcolor{TabGroup}\textbf{Height-aware blockage}} \\
\hspace{0.6em}Alarms raised (112 live scenes) & \textcolor{black!55}{$\downarrow$} & \textbf{1} & 60 & \hspace{0.6em}Divergences (15 healthy scenes) & \textcolor{black!55}{$\downarrow$} & \textbf{34} & 60 \\
\hspace{0.6em}False alarms among them & \textcolor{black!55}{$\downarrow$} & \textbf{0} & 59 & \hspace{0.6em}In-range coverage, Town10 (\%) & \textcolor{black!55}{$\uparrow$} & \textbf{100.0} & 47.7 \\
\multicolumn{4}{l}{\cellcolor{TabGroup}\textbf{Feasibility gate}} & \multicolumn{4}{l}{\cellcolor{TabGroup}\textbf{Scene planner (multi-approach junctions)}} \\
\hspace{0.6em}IMA admissions per 150 executions & \textcolor{black!55}{$\uparrow$} & \textbf{29.0} & 15.3 & \hspace{0.6em}Conflict gap, ground-truth ego (m) & \textcolor{black!55}{$\downarrow$} & \textbf{10.8} & 49.8 \\
\hspace{0.6em}Seeds with zero admissions (of 3) & \textcolor{black!55}{$\downarrow$} & \textbf{0} & 1 & \hspace{0.6em}Conflict gap, message-driven ego (m) & \textcolor{black!55}{$\downarrow$} & \textbf{2.4} & 5.7 \\
\multicolumn{4}{l}{\cellcolor{TabGroup}\textbf{Launch conditioning}} & \multicolumn{4}{l}{\cellcolor{TabGroup}\textbf{Prompt reflection}} \\
\hspace{0.6em}Opening cross-track error, median (m) & \textcolor{black!55}{$\downarrow$} & \textbf{0.06} & 1.22 & \hspace{0.6em}Non-admitted prompts rescued (of 43) & \textcolor{black!55}{$\uparrow$} & \textbf{8} & 0 \\
\hspace{0.6em}Opening cross-track error, worst (m) & \textcolor{black!55}{$\downarrow$} & \textbf{2.26} & 6.63 & \hspace{0.6em}Scenes admitted beyond search reach & \textcolor{black!55}{$\uparrow$} & \textbf{2} & 0 \\
\hspace{0.6em}Spurious ground-truth collisions & \textcolor{black!55}{$\downarrow$} & \textbf{0} & 1 &  & & \cellcolor{white} &  \\
\bottomrule
\end{tabularx}
\endgroup
\end{table}

Several findings are notable. First, the ground-truth rollout is essential for causal attribution. Among the 112 live scenes, 60 contain at least one decision made under a divergent estimated state. A divergence-only monitor would therefore raise 60 alarms, whereas the paired rollout identifies only one communication-induced failure, avoiding widespread false attribution. Second, the channel sensitivity study shows that building blockage is the dominant communication parameter. On Town10, reducing blockage loss from 5 to 1\,dB/m increases in-range road coverage from 47.7\% to 95.8\%, while changing transmit power by $\pm 10$\,dB shifts coverage by less than four percentage points. Height-aware blockage also avoids the systematic over-attenuation produced by a height-blind model. Third, the feasibility gate and scene planner matter primarily when their corresponding failure modes are present. Without the feasibility gate, infeasible mutants enter the search and reduce IMA admissions from 29.0 to 15.3 per 150-execution ablation budget, averaged over three seeds, while RLVW is largely unaffected because its mutations remain feasible. Similarly, the scene planner has little effect on simple junctions but becomes critical at multi-approach intersections, where heuristic grounding can eliminate the intended conflict. Finally, prompt reflection extends the refinement space beyond bounded parameter search. It rescues eight non-admitted prompts, including two cases that search alone cannot reach. When refinement is unsuccessful, reflection does not force an admission; instead, it identifies why the scenario remains ineffective or non-attributable.

\FloatBarrier
\section{Conclusion}
\label{sec:conclusion}
\label{sec:limits}
\label{sec:limitations}

This paper presented \sys{}, a language-driven, adaptive framework for automated scenario synthesis and safety testing of vehicle-to-everything (V2X) applications. By integrating large language models with domain knowledge, physical traffic simulation, and communication modeling, \sys{} transforms natural-language test descriptions into executable scenarios that capture the interactions between vehicle behavior and V2X information. Unlike conventional scenario-generation approaches that primarily focus on physical traffic conditions, \sys{} explicitly examines how communication degradation propagates through application decisions to affect driving safety. A paired execution mechanism distinguishes communication-induced failures from unsafe traffic conditions, while execution-guided parameter search and LLM self-reflection adaptively refine ineffective scenarios and identify minimal reproducible failures.

Experiments across diverse CARLA environments and V2X safety applications demonstrated the effectiveness of the proposed framework. \sys{} reliably generated executable scenarios from natural-language descriptions while avoiding spurious failure attribution under healthy communication. The paired execution mechanism further demonstrated the importance of causal verification, as discrepancies between received information and physical ground truth do not necessarily lead to unsafe outcomes. Adaptive refinement substantially improved the discovery of communication-induced failures by correcting ineffective fault configurations and revising test descriptions based on simulation evidence. Additional experiments highlighted how communication coverage, information availability, and cooperation among connected vehicles influence safety and driving behavior. Collectively, these findings demonstrate the value of integrating language-driven scenario generation, causal verification, and adaptive refinement into a unified V2X testing framework.

Several limitations remain. First, the communication model and fault configurations rely on representative parameters rather than comprehensive measurements from real-world deployments. Second, the evaluated safety applications employ simplified, deterministic decision logic that may not fully capture the complexity of production V2X systems. Third, although the main failure mechanisms were validated in CARLA, large-scale adaptive search was conducted using a simplified backend, and its effectiveness under fully integrated simulation remains to be established. Finally, the current framework assumes fixed-time traffic signals and does not explicitly distinguish communication errors from changes in signal timing caused by adaptive traffic control.

Future research will focus on improving the realism, scalability, and practical applicability of \sys{}. Integrating field-measured communication characteristics and more advanced V2X application logic will enable more representative evaluations of real-world connected vehicle systems. Extending adaptive refinement to fully integrated CARLA simulations and more complex traffic environments will further assess the framework's generalizability. Additional directions include supporting actuated signal control, expanding the range of communication and behavioral uncertainties, and using generated failure evidence to inform communication infrastructure design and V2X safety improvements. Ultimately, \sys{} provides a foundation for moving beyond manually specified test cases toward automated, evidence-driven safety evaluation of connected vehicle systems.

\FloatBarrier
\bibliographystyle{unsrtnat}
\bibliography{refs}

@inproceedings{zhang2024chatscene,
  author    = {Zhang, Jiawei and Xu, Chejian and Li, Bo},
  title     = {{ChatScene}: Knowledge-Enabled Safety-Critical Scenario Generation for Autonomous Vehicles},
  booktitle = {Proceedings of the IEEE/CVF Conference on Computer Vision and Pattern Recognition (CVPR)},
  pages     = {15459--15469},
  year      = {2024}
}

@inproceedings{ding2024realgen,
  author    = {Ding, Wenhao and Cao, Yulong and Zhao, Ding and Xiao, Chaowei and Pavone, Marco},
  title     = {{RealGen}: Retrieval Augmented Generation for Controllable Traffic Scenarios},
  booktitle = {Computer Vision -- ECCV 2024},
  series    = {Lecture Notes in Computer Science},
  publisher = {Springer},
  pages     = {93--110},
  year      = {2024},
  doi       = {10.1007/978-3-031-73033-7_6}
}

@inproceedings{zhang2023cat,
  author    = {Zhang, Linrui and Peng, Zhenghao and Li, Quanyi and Zhou, Bolei},
  title     = {{CAT}: Closed-loop Adversarial Training for Safe End-to-End Driving},
  booktitle = {Proceedings of the 7th Conference on Robot Learning (CoRL)},
  series    = {Proceedings of Machine Learning Research},
  volume    = {229},
  pages     = {2357--2372},
  year      = {2023},
  publisher = {PMLR}
}

@article{fremont2023scenic,
  author  = {Fremont, Daniel J. and Kim, Edward and Dreossi, Tommaso and Ghosh, Shromona and Yue, Xiangyu and Sangiovanni-Vincentelli, Alberto L. and Seshia, Sanjit A.},
  title   = {Scenic: A Language for Scenario Specification and Data Generation},
  journal = {Machine Learning},
  volume  = {112},
  number  = {10},
  pages   = {3805--3849},
  year    = {2023},
  doi     = {10.1007/s10994-021-06120-5}
}

@inproceedings{xu2022safebench,
  author    = {Xu, Chejian and Ding, Wenhao and Lyu, Weijie and Liu, Zuxin and Wang, Shuai and He, Yihan and Hu, Hanjiang and Zhao, Ding and Li, Bo},
  title     = {{SafeBench}: A Benchmarking Platform for Safety Evaluation of Autonomous Vehicles},
  booktitle = {Advances in Neural Information Processing Systems (NeurIPS), Datasets and Benchmarks Track},
  volume    = {35},
  pages     = {25667--25682},
  year      = {2022},
  doi       = {10.52202/068431-1861}
}

@inproceedings{ruan2025tts,
  author    = {Ruan, Bo-Kai and Tsui, Hao-Tang and Li, Yung-Hui and Shuai, Hong-Han},
  title     = {Traffic Scene Generation from Natural Language Description for Autonomous Vehicles with Large Language Model},
  booktitle = {Proceedings of the IEEE/CVF Conference on Computer Vision and Pattern Recognition (CVPR) Workshops},
  pages     = {935--944},
  year      = {2026}
}

@article{you2024v2x,
  title={{V2X-VLM}: End-to-End {V2X} Cooperative Autonomous Driving Through Large Vision-Language Models},
  author={You, Junwei and Jiang, Zhuoyu and Huang, Zilin and Shi, Haotian and Gan, Rui and Wu, Keshu and Cheng, Xi and Li, Xiaopeng and Ran, Bin},
  journal={Transportation Research Part C: Emerging Technologies},
  volume={183},
  pages={105457},
  year={2026},
  publisher={Elsevier},
  doi={10.1016/j.trc.2025.105457}
}

@article{wu2025v2x,
  title={{V2X-LLM}: Enhancing {V2X} Integration and Understanding in Connected Vehicle Corridors},
  author={Wu, Keshu and Li, Pei and Zhou, Yang and Gan, Rui and You, Junwei and Cheng, Yang and Zhu, Jingwen and Parker, Steven T. and Ran, Bin and Noyce, David A. and Tu, Zhengzhong},
  journal={arXiv preprint arXiv:2503.02239},
  year={2025}
}

@article{li2023does,
  author  = {Li, Pei and Wu, Keshu and Cheng, Yang and Parker, Steven T. and Noyce, David A.},
  title   = {How Does {C-V2X} Perform in Urban Environments? {Results} From Real-World Experiments on Urban Arterials},
  journal = {IEEE Transactions on Intelligent Vehicles},
  volume  = {9},
  number  = {1},
  pages   = {2520--2530},
  year    = {2024},
  doi     = {10.1109/TIV.2023.3326735}
}

@article{wu2025digital,
  author  = {Wu, Keshu and Li, Pei and Cheng, Yang and Parker, Steven T. and Ran, Bin and Noyce, David A. and Ye, Xinyue},
  title   = {A Digital Twin Framework for Physical-Virtual Integration in {V2X}-Enabled Connected Vehicle Corridors},
  journal = {IEEE Transactions on Intelligent Transportation Systems},
  volume  = {26},
  number  = {9},
  pages   = {14407--14420},
  year    = {2025},
  doi     = {10.1109/TITS.2025.3564351}
}

@article{gao2025airv2x,
  title={{AirV2X}: Unified Air-Ground Vehicle-to-Everything Collaboration},
  author={Gao, Xiangbo and Wu, Yuheng and Yang, Fengze and Luo, Xuewen and Wu, Keshu and Chen, Xinghao and Wang, Yuping and Liu, Chenxi and Zhou, Yang and Tu, Zhengzhong},
  journal={arXiv preprint arXiv:2506.19283},
  year={2025}
}

@article{wang2025generative,
  title={Generative {AI} for Autonomous Driving: Frontiers and Opportunities},
  author={Wang, Yuping and Xing, Shuo and Cui, Can and Li, Renjie and Hua, Hongyuan and Tian, Kexin and Mo, Zhaobin and Gao, Xiangbo and Pavone, Marco and Zhou, Yang and Li, Jiachen and Tu, Zhengzhong},
  journal={ACM Computing Surveys},
  year={2026},
  publisher={ACM},
  doi={10.1145/3838726}
}

@inproceedings{wu2023development,
  author    = {Wu, Keshu and Cheng, Yang and Parker, Steven T. and Ran, Bin and Noyce, David A.},
  title     = {Development of the Data Pipeline for a Connected Vehicle Corridor},
  booktitle = {International Conference on Transportation and Development 2023},
  publisher = {ASCE},
  pages     = {218--230},
  year      = {2023},
  doi       = {10.1061/9780784484876.020}
}

@inproceedings{yu2023v2x,
  author    = {Yu, Haibao and Yang, Wenxian and Ruan, Hongzhi and Yang, Zhenwei and Tang, Yingjuan and Gao, Xu and Hao, Xin and Shi, Yifeng and Pan, Yifeng and Sun, Ning and others},
  title     = {{V2X-Seq}: A Large-Scale Sequential Dataset for Vehicle-Infrastructure Cooperative Perception and Forecasting},
  booktitle = {Proceedings of the IEEE/CVF Conference on Computer Vision and Pattern Recognition (CVPR)},
  pages     = {5486--5495},
  year      = {2023}
}

@article{maglogiannis2021experimental,
  author  = {Maglogiannis, Vasilis and Naudts, Dries and Hadiwardoyo, Seilendria and van den Akker, Daniel and Marquez-Barja, Johann and Moerman, Ingrid},
  title   = {Experimental {V2X} Evaluation for {C-V2X} and {ITS-G5} Technologies in a Real-Life Highway Environment},
  journal = {IEEE Transactions on Network and Service Management},
  volume  = {19},
  number  = {2},
  pages   = {1521--1538},
  year    = {2022},
  doi     = {10.1109/TNSM.2021.3129348}
}

@article{liao2024digital,
  author  = {Liao, Xiwen and Leng, Supeng and Sun, Yao and Zhang, Ke and Imran, Muhammad Ali},
  title   = {A Digital-Twin-Based Traffic Guidance Scheme for Autonomous Driving},
  journal = {IEEE Internet of Things Journal},
  volume  = {11},
  number  = {22},
  pages   = {36829--36840},
  year    = {2024},
  doi     = {10.1109/JIOT.2024.3429534}
}

@inproceedings{ding2020learning,
  author    = {Ding, Wenhao and Chen, Baiming and Xu, Minjun and Zhao, Ding},
  title     = {Learning to Collide: An Adaptive Safety-Critical Scenarios Generating Method},
  booktitle = {2020 IEEE/RSJ International Conference on Intelligent Robots and Systems (IROS)},
  pages     = {2243--2250},
  year      = {2020},
  doi       = {10.1109/IROS45743.2020.9340696}
}

@inproceedings{wang2021advsim,
  title={Advsim: Generating safety-critical scenarios for self-driving vehicles},
  author={Wang, Jingkang and Pun, Ava and Tu, James and Manivasagam, Sivabalan and Sadat, Abbas and Casas, Sergio and Ren, Mengye and Urtasun, Raquel},
  booktitle={Proceedings of the IEEE/CVF Conference on Computer Vision and Pattern Recognition},
  pages={9909--9918},
  year={2021}
}

@inproceedings{dosovitskiy2017carla,
  author    = {Dosovitskiy, Alexey and Ros, German and Codevilla, Felipe and Lopez, Antonio and Koltun, Vladlen},
  title     = {{CARLA}: An Open Urban Driving Simulator},
  booktitle = {Proceedings of the 1st Annual Conference on Robot Learning (CoRL)},
  series    = {Proceedings of Machine Learning Research},
  volume    = {78},
  pages     = {1--16},
  year      = {2017},
  publisher = {PMLR}
}

@article{li2022metadrive,
  author  = {Li, Quanyi and Peng, Zhenghao and Feng, Lan and Zhang, Qihang and Xue, Zhenghai and Zhou, Bolei},
  title   = {{MetaDrive}: Composing Diverse Driving Scenarios for Generalizable Reinforcement Learning},
  journal = {IEEE Transactions on Pattern Analysis and Machine Intelligence},
  volume  = {45},
  number  = {3},
  pages   = {3461--3475},
  year    = {2023},
  doi     = {10.1109/TPAMI.2022.3190471}
}

@article{cai2023,
  author  = {Cai, Guoqiang and Fan, Bo and Dong, Yiwei and Li, Tongfei and Wu, Yuan and Zhang, Yan},
  title   = {Task-Efficiency Oriented {V2X} Communications: Digital Twin Meets Mobile Edge Computing},
  journal = {IEEE Wireless Communications},
  volume  = {31},
  number  = {2},
  pages   = {149--155},
  year    = {2024},
  doi     = {10.1109/MWC.012.2200465}
}

@inproceedings{zelenbaba2022wireless,
  author    = {Zelenbaba, Stefan and Rainer, Benjamin and Hofer, Markus and Zemen, Thomas},
  title     = {Wireless Digital Twin for Assessing the Reliability of Vehicular Communication Links},
  booktitle = {2022 IEEE Globecom Workshops (GC Wkshps)},
  pages     = {1034--1039},
  year      = {2022},
  doi       = {10.1109/GCWkshps56602.2022.10008559}
}

@article{liu2022application,
  author  = {Liu, Qian and Qi, Xiaoqing and Liu, Shanshan and Cheng, Xiuju and Ke, Xin and Wang, Feng},
  title   = {Application of Lightweight Digital Twin System in Intelligent Transportation},
  journal = {IEEE Journal of Radio Frequency Identification},
  volume  = {6},
  pages   = {729--732},
  year    = {2022},
  doi     = {10.1109/JRFID.2022.3212169}
}

@inproceedings{sheng2025talk2traffic,
  author    = {Sheng, Zihao and Huang, Zilin and Qu, Yansong and Leng, Yue and Chen, Sikai},
  title     = {{Talk2Traffic}: Interactive and Editable Traffic Scenario Generation for Autonomous Driving with Multimodal Large Language Model},
  booktitle = {Proceedings of the IEEE/CVF Conference on Computer Vision and Pattern Recognition Workshops (CVPRW)},
  pages     = {3788--3797},
  year      = {2025},
  doi       = {10.1109/CVPRW67362.2025.00364}
}

@article{zhang2025virtual,
  author  = {Zhang, Hao and Yue, Ximin and Tian, Kexin and Li, Sixu and Wu, Keshu and Li, Zihao and Lord, Dominique and Zhou, Yang},
  title   = {Virtual Roads, Smarter Safety: A Digital Twin Framework for Mixed Autonomous Traffic Safety Analysis},
  journal = {IEEE Internet of Things Journal},
  volume  = {13},
  number  = {4},
  pages   = {6045--6058},
  year    = {2026},
  doi     = {10.1109/JIOT.2025.3637688}
}

@misc{sae-j2735_2016,
  author       = {{SAE International}},
  title        = {{V2X} Communications Message Set Dictionary},
  howpublished = {SAE Standard J2735\_202409},
  month        = sep,
  year         = {2024},
  doi          = {10.4271/J2735_202409}
}

@misc{gpt4o,
  author        = {{OpenAI}},
  title         = {{GPT-4o} System Card},
  year          = {2024},
  howpublished  = {arXiv preprint arXiv:2410.21276},
  doi           = {10.48550/arXiv.2410.21276}
}

@misc{singh2025openai,
  author        = {Singh, Aaditya and Fry, Adam and Perelman, Adam and Tart, Adam and Ganesh, Adi and El-Kishky, Ahmed and McLaughlin, Aidan and Low, Aiden and Ostrow, AJ and Ananthram, Akhila and others},
  title         = {{OpenAI} {GPT-5} System Card},
  year          = {2025},
  howpublished  = {arXiv preprint arXiv:2601.03267},
  doi           = {10.48550/arXiv.2601.03267}
}

@article{YI2024100126,
  author  = {Yi, Ran and Yao, Yifan and Pu, Fan and Zhou, Yang and Wang, Xin},
  title   = {Cooperative {CAV} mandatory lane-change control enabled by {V2I}},
  journal = {Communications in Transportation Research},
  volume  = {4},
  pages   = {100126},
  year    = {2024},
  doi     = {10.1016/j.commtr.2024.100126}
}

@article{yang2026edge,
  author  = {Yang, Fengze and Yu, Bo and Zhou, Yang and Luo, Xuewen and Tu, Zhengzhong and Liu, Chenxi},
  title   = {Edge-based multimodal sensor data fusion with Vision-Language-Action ({VLA}) model for real-time autonomous vehicle accident avoidance},
  journal = {Engineering Applications of Artificial Intelligence},
  volume  = {180},
  pages   = {115186},
  year    = {2026},
  doi     = {10.1016/j.engappai.2026.115186}
}

@inproceedings{li2026cats,
  author    = {Li, Hangyu and Cao, Bofeng and Liang, Zhaohui and Li, Wuzhen and Oh, Juyoung and Chen, Yuxuan and Liang, Shixiao and Zhou, Hang and Ma, Chengyuan and Liu, Jiaxi and others},
  title     = {{CATS-V2V}: A Real-World Vehicle-to-Vehicle Cooperative Perception Dataset with Complex Adverse Traffic Scenarios},
  booktitle = {Proceedings of the IEEE/CVF Conference on Computer Vision and Pattern Recognition (CVPR) Findings},
  pages     = {2294--2303},
  year      = {2026},
  doi       = {10.1109/CVPRF71909.2026.00224}
}

@article{LI2026105759,
  author  = {Li, Shuyang and Ma, Meng and Azfar, Talha and Ke, Ruimin},
  title   = {{ChatSUMO} Agent: An {LLM}-Based Agent for Conversational Traffic Simulation in {SUMO}},
  journal = {Transportation Research Part C: Emerging Technologies},
  volume  = {190},
  pages   = {105759},
  year    = {2026},
  doi     = {10.1016/j.trc.2026.105759}
}

@article{feng2023dense,
  author  = {Feng, Shuo and Sun, Haowei and Yan, Xintao and Zhu, Haojie and Zou, Zhengxia and Shen, Shengyin and Liu, Henry X.},
  title   = {Dense reinforcement learning for safety validation of autonomous vehicles},
  journal = {Nature},
  volume  = {615},
  number  = {7953},
  pages   = {620--627},
  year    = {2023},
  doi     = {10.1038/s41586-023-05732-2}
}

@article{feng2021intelligent,
  author  = {Feng, Shuo and Yan, Xintao and Sun, Haowei and Feng, Yiheng and Liu, Henry X.},
  title   = {Intelligent driving intelligence test for autonomous vehicles with naturalistic and adversarial environment},
  journal = {Nature Communications},
  volume  = {12},
  number  = {1},
  pages   = {748},
  year    = {2021},
  doi     = {10.1038/s41467-021-21007-8}
}

@article{mei2025llmattacker,
  author  = {Mei, Yuewen and Nie, Tong and Sun, Jian and Tian, Ye},
  title   = {{LLM-Attacker}: Enhancing Closed-Loop Adversarial Scenario Generation for Autonomous Driving With Large Language Models},
  journal = {IEEE Transactions on Intelligent Transportation Systems},
  volume  = {26},
  number  = {10},
  pages   = {15068--15076},
  year    = {2025},
  doi     = {10.1109/TITS.2025.3578383}
}

@inproceedings{tang2024legend,
  author    = {Tang, Shuncheng and Zhang, Zhenya and Zhou, Jixiang and Lei, Lei and Zhou, Yuan and Xue, Yinxing},
  title     = {{LeGEND}: A Top-Down Approach to Scenario Generation of Autonomous Driving Systems Assisted by Large Language Models},
  booktitle = {Proceedings of the 39th IEEE/ACM International Conference on Automated Software Engineering (ASE)},
  pages     = {1497--1508},
  publisher = {ACM},
  year      = {2024},
  doi       = {10.1145/3691620.3695520}
}

@article{tu2025trafficcomposer,
  author  = {Tu, Zhi and Niu, Liangkun and Fan, Wei and Zhang, Tianyi},
  title   = {Multi-modal Traffic Scenario Generation for Autonomous Driving System Testing},
  journal = {Proceedings of the ACM on Software Engineering},
  volume  = {2},
  number  = {FSE},
  pages   = {1733--1756},
  year    = {2025},
  doi     = {10.1145/3729348}
}

@article{peng2026ldscene,
  author  = {Peng, Mingxing and Xie, Yuting and Guo, Xusen and Yao, Ruoyu and Yang, Hai and Ma, Jun},
  title   = {{LD-Scene}: {LLM}-Guided Diffusion for Controllable Generation of Adversarial Safety-Critical Driving Scenarios},
  journal = {Transportation Research Part C: Emerging Technologies},
  volume  = {188},
  pages   = {105694},
  year    = {2026},
  doi     = {10.1016/j.trc.2026.105694}
}

@article{cai2026text2scenario,
  author  = {Cai, Xuan and Bai, Xuesong and Cui, Zhiyong and Xie, Danmu and Fu, Daocheng and Yu, Haiyang and Ren, Yilong},
  title   = {{Text2Scenario}: Text-Driven Scenario Generation for Autonomous Driving Test},
  journal = {Automotive Innovation},
  volume  = {9},
  number  = {1},
  pages   = {102--127},
  year    = {2026},
  doi     = {10.1007/s42154-025-00374-8}
}

@inproceedings{guo2024cootest,
  author    = {Guo, An and Gao, Xinyu and Chen, Zhenyu and Xiao, Yuan and Liu, Jiakai and Ge, Xiuting and Sun, Weisong and Fang, Chunrong},
  title     = {{CooTest}: An Automated Testing Approach for {V2X} Communication Systems},
  booktitle = {Proceedings of the 33rd ACM SIGSOFT International Symposium on Software Testing and Analysis (ISSTA)},
  pages     = {1453--1465},
  publisher = {ACM},
  year      = {2024},
  doi       = {10.1145/3650212.3680373}
}

@inproceedings{zhang2024fabrication,
  author    = {Zhang, Qingzhao and Jin, Shuowei and Zhu, Ruiyang and Sun, Jiachen and Zhang, Xumiao and Chen, Qi Alfred and Mao, Z. Morley},
  title     = {On Data Fabrication in Collaborative Vehicular Perception: Attacks and Countermeasures},
  booktitle = {Proceedings of the 33rd USENIX Security Symposium (USENIX Security)},
  pages     = {6309--6326},
  publisher = {USENIX Association},
  year      = {2024}
}

@article{xu2023opencda,
  author  = {Xu, Runsheng and Xiang, Hao and Han, Xu and Xia, Xin and Meng, Zonglin and Chen, Chia-Ju and Correa-Jullian, Camila and Ma, Jiaqi},
  title   = {The {OpenCDA} Open-Source Ecosystem for Cooperative Driving Automation Research},
  journal = {IEEE Transactions on Intelligent Vehicles},
  volume  = {8},
  number  = {4},
  pages   = {2698--2711},
  year    = {2023},
  doi     = {10.1109/TIV.2023.3244948}
}

@inproceedings{grimm2024cosimulate,
  author    = {Grimm, Daniel and Schindewolf, Marc and Kraus, David and Sax, Eric},
  title     = {Co-simulate no more: The {CARLA} {V2X} Sensor},
  booktitle = {2024 IEEE Intelligent Vehicles Symposium (IV)},
  pages     = {2429--2436},
  year      = {2024},
  doi       = {10.1109/IV55156.2024.10588488}
}

@inproceedings{kaul2012realtime,
  author    = {Kaul, Sanjit and Yates, Roy and Gruteser, Marco},
  title     = {Real-time status: How often should one update?},
  booktitle = {Proceedings of IEEE INFOCOM},
  pages     = {2731--2735},
  year      = {2012},
  doi       = {10.1109/INFCOM.2012.6195689}
}

@inproceedings{shinn2023reflexion,
  author    = {Shinn, Noah and Cassano, Federico and Gopinath, Ashwin and Narasimhan, Karthik and Yao, Shunyu},
  title     = {Reflexion: Language Agents with Verbal Reinforcement Learning},
  booktitle = {Advances in Neural Information Processing Systems (NeurIPS)},
  volume    = {36},
  pages     = {8634--8652},
  year      = {2023},
  doi       = {10.52202/075280-0377}
}

@inproceedings{madaan2023selfrefine,
  author    = {Madaan, Aman and Tandon, Niket and Gupta, Prakhar and Hallinan, Skyler and Gao, Luyu and Wiegreffe, Sarah and Alon, Uri and Dziri, Nouha and Prabhumoye, Shrimai and Yang, Yiming and others},
  title     = {Self-Refine: Iterative Refinement with Self-Feedback},
  booktitle = {Advances in Neural Information Processing Systems (NeurIPS)},
  volume    = {36},
  pages     = {46534--46594},
  year      = {2023},
  doi       = {10.52202/075280-2019}
}

@misc{sae-j2945-1_2020,
  author       = {{SAE International}},
  title        = {On-Board System Requirements for {V2V} Safety Communications},
  howpublished = {SAE Standard J2945/1\_202004},
  month        = apr,
  year         = {2020},
  doi          = {10.4271/J2945/1_202004}
}

@article{wu2026aurora,
  title={AURORA: A Natural Language-Driven Agentic Framework for Understanding, Reasoning, and Orchestrating Reliable Air-Ground Co-Simulation},
  author={Wu, Keshu and Zhang, Hao and Gan, Rui and Gao, Xiangbo and Li, Xiaopeng and Tu, Zhengzhong and Zhou, Yang},
  journal={arXiv preprint arXiv:2609.19527},
  year={2026}
}

@article{wu2025ai2,
  title={{AI2-Active Safety}: {AI}-enabled Interaction-aware Active Safety Analysis with Vehicle Dynamics},
  author={Wu, Keshu and Li, Zihao and Li, Sixu and Ye, Xinyue and Lord, Dominique and Zhou, Yang},
  journal={arXiv preprint arXiv:2505.00322},
  year={2025}
}

@article{wu2026knowledge,
  title={Knowledge Is Not Static: Order-Aware Hypergraph {RAG} for Language Models},
  author={Wu, Keshu and Kuai, Chenchen and Li, Zihao and Jiang, Jiwan and Shen, Shiyu and Wang, Shian and Hu, Chan-Wei and Tu, Zhengzhong and Zhou, Yang},
  journal={arXiv preprint arXiv:2604.12185},
  year={2026}
}

@inproceedings{mei2025seeking,
  author    = {Mei, Yuewen and Nie, Tong and Sun, Jian and Tian, Ye},
  title     = {Seeking to Collide: Online Safety-Critical Scenario Generation for Autonomous Driving with Retrieval Augmented Large Language Models},
  booktitle = {2025 IEEE 28th International Conference on Intelligent Transportation Systems (ITSC)},
  pages     = {943--950},
  year      = {2025},
  doi       = {10.1109/ITSC60802.2025.11423790}
}

@inproceedings{gao2025words,
  author    = {Gao, Yuan and Piccinini, Mattia and Moller, Korbinian and Alanwar, Amr and Betz, Johannes},
  title     = {From Words to Collisions: {LLM}-Guided Evaluation and Adversarial Generation of Safety-Critical Driving Scenarios},
  booktitle = {2025 IEEE 28th International Conference on Intelligent Transportation Systems (ITSC)},
  pages     = {2134--2141},
  year      = {2025},
  doi       = {10.1109/ITSC60802.2025.11423486}
}

@inproceedings{gao2025laser,
  author    = {Gao, Hao and Wang, Jingyue and Fang, Wenyang and Xu, Jingwei and Huang, Yunpeng and Chen, Taolue and Ma, Xiaoxing},
  title     = {{LASER}: Script Execution by Autonomous Agents for On-demand Traffic Simulation},
  booktitle = {Proceedings of the 16th International Conference on Internetware},
  publisher = {ACM},
  pages     = {84--95},
  year      = {2025},
  doi       = {10.1145/3755881.3755905}
}

@article{zhao2026survey,
  author  = {Zhao, Yongqi and Zhou, Ji and Bi, Dong and Mihalj, Tomislav and Hu, Jia and Eichberger, Arno},
  title   = {A Survey on the Application of Large Language Models in Scenario-Based Testing of Automated Driving Systems},
  journal = {IEEE Transactions on Intelligent Transportation Systems},
  volume  = {27},
  number  = {5},
  pages   = {5001--5023},
  year    = {2026},
  doi     = {10.1109/TITS.2026.3651004}
}

@article{wu2025hypergraph,
  title={Hypergraph-based motion generation with multi-modal interaction relational reasoning},
  author={Wu, Keshu and Zhou, Yang and Shi, Haotian and Lord, Dominique and Ran, Bin and Ye, Xinyue},
  journal={Transportation Research Part C: Emerging Technologies},
  volume={180},
  pages={105349},
  year={2025},
  publisher={Elsevier},
  doi={10.1016/j.trc.2025.105349}
}

@inproceedings{tan2025promptable,
  title={Promptable Closed-loop Traffic Simulation},
  author={Tan, Shuhan and Ivanovic, Boris and Chen, Yuxiao and Li, Boyi and Weng, Xinshuo and Cao, Yulong and Kraehenbuehl, Philipp and Pavone, Marco},
  booktitle={Proceedings of The 8th Conference on Robot Learning},
  series={Proceedings of Machine Learning Research},
  volume={270},
  pages={5087--5105},
  publisher={PMLR},
  year={2025}
}

@inproceedings{elmaaroufi2024scenicnl,
  title={{ScenicNL}: Generating Probabilistic Scenario Programs from Natural Language},
  author={Elmaaroufi, Karim and Shanker, Devan and Cismaru, Ana and Vazquez-Chanlatte, Marcell and Sangiovanni-Vincentelli, Alberto L. and Zaharia, Matei and Seshia, Sanjit A.},
  booktitle={Conference on Language Modeling (COLM)},
  year={2024}
}

@article{wang2026learning,
  title={Learning from Risk: {LLM}-Guided Generation of Safety-Critical Scenarios with Prior Knowledge},
  author={Wang, Yuhang and Huang, Heye and Xu, Zhenhua and Sun, Kailai and Guo, Baoshen and Zhao, Jinhua},
  journal={Transportation Research Part E: Logistics and Transportation Review},
  volume={215},
  pages={105110},
  year={2026},
  doi={10.1016/j.tre.2026.105110}
}

@article{zhao2026sg,
  title={{SG-CADVLM}: A Context-Aware Decoding Powered Vision-Language Model for Safety-Critical Scenario Generation in Autonomous Driving},
  author={Zhao, Hongyi and Wang, Shuo and He, Qijie and Pu, Ziyuan},
  journal={Transportmetrica A: Transport Science},
  pages={1--27},
  year={2026},
  doi={10.1080/23249935.2026.2666104}
}

@inproceedings{zhou2025decoupled,
  title={Decoupled Diffusion Sparks Adaptive Scene Generation},
  author={Zhou, Yunsong and Ye, Naisheng and Ljungbergh, William and Li, Tianyu and Yang, Jiazhi and Yang, Zetong and Zhu, Hongzi and Petersson, Christoffer and Li, Hongyang},
  booktitle={Proceedings of the IEEE/CVF International Conference on Computer Vision (ICCV)},
  pages={27760--27770},
  year={2025},
  doi={10.1109/ICCV51701.2025.02577}
}

@inproceedings{guo2024sovar,
  title={{SoVAR}: Build Generalizable Scenarios from Accident Reports for Autonomous Driving Testing},
  author={Guo, An and Zhou, Yuan and Tian, Haoxiang and Fang, Chunrong and Sun, Yunjian and Sun, Weisong and Gao, Xinyu and Luu, Anh Tuan and Liu, Yang and Chen, Zhenyu},
  booktitle={Proceedings of the 39th IEEE/ACM International Conference on Automated Software Engineering (ASE)},
  pages={268--280},
  year={2024},
  doi={10.1145/3691620.3695037}
}

@inproceedings{liu2025colmdriver,
  title={{CoLMDriver}: {LLM}-based Negotiation Benefits Cooperative Autonomous Driving},
  author={Liu, Changxing and Liu, Genjia and Wang, Zijun and Yang, Jinchang and Chen, Siheng},
  booktitle={Proceedings of the IEEE/CVF International Conference on Computer Vision (ICCV)},
  pages={25951--25960},
  year={2025},
  doi={10.1109/ICCV51701.2025.02407}
}

@inproceedings{chiu2026v2v,
  title={{V2V-LLM}: Vehicle-to-Vehicle Cooperative Autonomous Driving with Multimodal Large Language Models},
  author={Chiu, Hsu-kuang and Hachiuma, Ryo and Wang, Chien-Yi and Smith, Stephen F. and Wang, Yu-Chiang Frank and Chen, Min-Hung},
  booktitle={2026 IEEE International Conference on Robotics and Automation (ICRA)},
  pages={12538--12545},
  year={2026},
  doi={10.1109/ICRA57385.2026.11696625}
}

@inproceedings{yu2025end,
  title={End-to-End Autonomous Driving Through {V2X} Cooperation},
  author={Yu, Haibao and Yang, Wenxian and Zhong, Jiaru and Yang, Zhenwei and Fan, Siqi and Luo, Ping and Nie, Zaiqing},
  booktitle={Proceedings of the AAAI Conference on Artificial Intelligence},
  volume={39},
  pages={9598--9606},
  year={2025},
  doi={10.1609/aaai.v39i9.33040}
}

@inproceedings{gao2025langcoop,
  title={{LangCoop}: Collaborative Driving with Language},
  author={Gao, Xiangbo and Wu, Yuheng and Wang, Rujia and Liu, Chenxi and Zhou, Yang and Tu, Zhengzhong},
  booktitle={Proceedings of the IEEE/CVF Conference on Computer Vision and Pattern Recognition Workshops (CVPRW)},
  pages={4226--4237},
  year={2025},
  doi={10.1109/CVPRW67362.2025.00406}
}

\clearpage
\appendix
\makeatletter
\let\lastv@seccntformat\@seccntformat
\renewcommand{\@seccntformat}[1]{\expandafter\ifx\csname c@#1\endcsname\c@section Appendix~\fi\lastv@seccntformat{#1}}
\makeatother

\section{Verifier Check Families}
\label{app:checks}
The analytic verifier of Module~I (Section~\ref{sec:parser}) screens every candidate specification before any simulator time is spent. Each check is decidable from the specification and the closed-form AoI relations alone, and returns a finding that names the check, its severity, the specification fields responsible, and a minimal repair, such as the breaching drop count $k^\ast$ of Eq.~(\ref{eq:kstar}) or a minimum episode duration. Severity determines how a finding is used:
\begin{itemize}
\item \textbf{Error} findings are blocking. During intent parsing they are returned to the LLM parser, which revises the specification; during specification-space search (Section~\ref{sec:refine}) a mutant with any error finding is counted as gated and is never executed.
\item \textbf{Warning} findings do not block execution but mark a scenario whose outcome is unlikely to be attributable or informative, for example because the pristine channel already exceeds an age budget.
\item \textbf{Note} findings record a property of the scenario, such as a deliberate silent-miss probe in which every conflict agent is unequipped.
\end{itemize}
Verifier diagnostics also enter the evidence digest used by prompt-space reflection. Table~\ref{tab:checks} lists the eleven check families. The conflict-timing check uses an arrival-overlap window of 8\,s, and the penetration check warns below a market penetration rate of 0.2.

\begin{table}[!ht]
\centering\footnotesize
\caption{Verifier check families, grouped by severity.}
\label{tab:checks}
\begingroup
\setlength{\aboverulesep}{0pt}
\setlength{\belowrulesep}{0pt}
\renewcommand{\arraystretch}{1.22}
\begin{tabularx}{\linewidth}{l >{\raggedright\arraybackslash}X}
\toprule
\textbf{Check} & \textbf{Condition rejected or flagged} \\
\midrule
\rowcolor{TabGroup}\multicolumn{2}{l}{\sevtag{SevError}{error}\enspace\textit{blocking; returned to the parser for repair}} \\
\hspace{0.5em}\code{no\_effective\_fault} & No fault can breach any application budget (a vacuous intent); suggestions name $k^\ast$ or a freeze/skew fault \\
\hspace{0.5em}\code{anchor\_beyond\_horizon} & The $k$-th anchored occurrence, or the ego decision point, falls after the horizon $T$ \\
\hspace{0.5em}\code{agent\_unreachable} & The conflict cannot form \\
\hspace{0.5em}\code{no\_locus} & Retrieval returns no locus; the message names the first filter that empties the candidate set \\
\rowcolor{TabGroup}\multicolumn{2}{l}{\sevtag{SevWarn}{warn}\enspace\textit{executed, but flagged as unlikely to be attributable}} \\
\hspace{0.5em}\code{fault\_below\_budget} & A particular fault is invisible to the safety criterion \\
\hspace{0.5em}\code{baseline\_blows\_budget} & The pristine channel already exceeds a budget, so failures are not attributable to faults \\
\hspace{0.5em}\code{conflict\_timing\_mismatch} & The arrival gap exceeds the overlap window \\
\hspace{0.5em}\code{no\_closing\_speed} & The FCW/EEBL lead vehicle is not closing \\
\hspace{0.5em}\code{mpr\_low\_unpinned} & Regimes mix under low market penetration \\
\rowcolor{TabGroup}\multicolumn{2}{l}{\sevtag{SevNote}{note}\enspace\textit{recorded with the scenario}} \\
\hspace{0.5em}\code{silent\_miss\_mode} & All conflict agents are pinned unequipped (structural probe) \\
\hspace{0.5em}\code{outside\_rsu\_range\_at\_spawn} & Reception begins mid-approach \\
\bottomrule
\end{tabularx}
\endgroup
\end{table}

\section{Channel Model Details}
\label{app:qual}
Every V2I and V2V message passes independently through the spatial channel of Module~III (Section~\ref{sec:grounding}); Table~\ref{tab:channel} lists its default parameters. For a transmitter--receiver pair at distance $d$, the received signal strength of Eq.~(\ref{eq:rssi}) is evaluated as
\begin{equation}
\mathrm{RSSI} = \max\!\Big(P_{\mathrm{tx}} - 10n\log_{10}\max(d,1\,\mathrm{m})
- B_{\mathrm{blk}},\; P_{\mathrm{floor}}\Big),
\label{eq:rssi-full}
\end{equation}
and a link with $d > R$ is assigned the noise floor.

\paragraph{Height-Aware Blockage}
The blockage term of Eq.~(\ref{eq:rssi}) is
\begin{equation}
B_{\mathrm{blk}} \;=\; \min\!\Big(
\eta \!\int_{\mathrm{path}}\!
\min\!\big(1,\, h_{\mathrm{exc}}(s)/h_{\mathrm{sat}}\big)\, ds,\;\; B_{\mathrm{NLOS}} \Big),
\label{eq:blockage}
\end{equation}
where $h_{\mathrm{exc}}(s)=\max\big(0,\,h_{\mathrm{b}}(s)-h_{\mathrm{ray}}(s)\big)$ is the height by which the building at path position $s$ rises above the sight line, and $h_{\mathrm{ray}}(s)$ falls linearly from the RSU antenna height $h_{\mathrm{RSU}}$ to the receiver antenna height $h_{\mathrm{OBU}}$. The integral is evaluated by sampling the ray at steps of $\Delta s$. Loss therefore accrues only where a building stands above the line of sight, grows with how far above it stands until it saturates at $h_{\mathrm{sat}}$, and is capped at $B_{\mathrm{NLOS}}$: however deep the obstruction, energy diffracted over rooflines and reflected from facades still reaches the receiver.

\paragraph{Latency and Delivery}
Latency grows linearly once the signal falls below a knee,
\begin{equation}
L = L_0 + \kappa\,\max\!\big(0,\; \mathrm{RSSI}_{\mathrm{knee}} - \mathrm{RSSI}\big),
\label{eq:latency}
\end{equation}
and per-message delivery is logistic in RSSI,
\begin{equation}
p_{\mathrm{del}} = \frac{1}{1+\exp\!\big((\mathrm{RSSI}_{50}-\mathrm{RSSI})/\nu\big)}.
\label{eq:delivery}
\end{equation}
Together, Eqs.~(\ref{eq:latency}) and (\ref{eq:delivery}) realize the mapping $g$ of Eq.~(\ref{eq:rssi}). A message is lost if its RSSI is at the noise floor or a Bernoulli draw with probability $1-p_{\mathrm{del}}$ fails; otherwise it is delivered after $L$. The semantic faults of Module~III act on top of this channel.

\begin{table}[!ht]
\centering\footnotesize
\caption{Default channel parameters. $^{\dagger}$The corridor study varies $R$ over 50, 100, and 150\,m.}
\label{tab:channel}
\begingroup
\setlength{\aboverulesep}{0pt}
\setlength{\belowrulesep}{0pt}
\renewcommand{\arraystretch}{1.22}
\begin{tabularx}{\linewidth}{>{\raggedright\arraybackslash}X c r @{\hspace{1.1em}{\color{black!25}\vrule width 0.4pt}\hspace{1.1em}} >{\raggedright\arraybackslash}X c r}
\toprule
\textbf{Parameter} & \textbf{Symbol} & \textbf{Value} & \textbf{Parameter} & \textbf{Symbol} & \textbf{Value} \\
\midrule
\multicolumn{3}{l}{\cellcolor{TabGroup}\textbf{Path loss}} & \multicolumn{3}{l}{\cellcolor{TabGroup}\textbf{Height-aware blockage}} \\
\hspace{0.5em}Transmit power & $P_{\mathrm{tx}}$ & 0\,dBm & \hspace{0.5em}Building loss rate & $\eta$ & 5\,dB/m \\
\hspace{0.5em}Path-loss exponent & $n$ & 2.0 & \hspace{0.5em}Ray sampling step & $\Delta s$ & 1\,m \\
\hspace{0.5em}Noise floor & $P_{\mathrm{floor}}$ & $-100$\,dBm & \hspace{0.5em}RSU antenna height & $h_{\mathrm{RSU}}$ & 6.5\,m \\
\hspace{0.5em}Communication radius & $R$ & 100\,m$^{\dagger}$ & \hspace{0.5em}Receiver antenna height & $h_{\mathrm{OBU}}$ & 1.5\,m \\
\multicolumn{3}{l}{\cellcolor{TabGroup}\textbf{Latency and delivery}} & \hspace{0.5em}Saturation excess height & $h_{\mathrm{sat}}$ & 10\,m \\
\hspace{0.5em}Base latency & $L_0$ & 10\,ms & \hspace{0.5em}NLOS loss cap & $B_{\mathrm{NLOS}}$ & 25\,dB \\
\hspace{0.5em}Latency slope & $\kappa$ & 0.2\,ms/dB & \multicolumn{3}{l}{\cellcolor{TabGroup}\textbf{Message rates}} \\
\hspace{0.5em}Latency knee & $\mathrm{RSSI}_{\mathrm{knee}}$ & $-70$\,dBm & \hspace{0.5em}SPaT / MAP / BSM & $1/T_s$ & 10 / 1 / 10\,Hz \\
\hspace{0.5em}Delivery midpoint & $\mathrm{RSSI}_{50}$ & $-90$\,dBm &  & &  \\
\hspace{0.5em}Delivery scale & $\nu$ & 5\,dB &  & &  \\
\bottomrule
\end{tabularx}
\endgroup
\end{table}

\section{Fault Operators}
\label{app:faults}
Each fault is a typed primitive that states what is corrupted (the target stream, SPaT, MAP, or BSM, and the field), when, and how much. Its activation window is $[t_a+o,\;t_a+o+\tau)$, where $t_a$ is the time of the $k$-th occurrence of an anchor event, $o$ is an offset that may be negative, and an unset duration $\tau$ keeps the fault active until the scenario ends. The anchor events are \code{sim\_start}, \code{green\_onset}, \code{amber\_onset}, \code{red\_onset}, \code{phase\_change}, \code{ego\_at\_decision\_point}, and \code{ego\_enters\_range}; signal anchors resolve against the bound signal group and kinematic anchors against the message-driven ego. A fault may be restricted to one sender, and each records the provenance of its magnitude (measured or representative); the experiments in this paper use representative magnitudes. Table~\ref{tab:faults} defines the seven operators together with the parameter ranges explored by specification-space search. Search additionally moves window offsets in steps of $\pm2.5$\,s within $[-10,10]$\,s, sets durations to $\{2,5,10\}$\,s or open, and changes the anchor event.

A \code{freeze} captures whatever the payload shows when its window opens. A feed that should appear stuck at green is therefore anchored before the amber onset (for example, \code{amber\_onset} with an offset of $-3$\,s or earlier); a window anchored at \code{red\_onset} captures yellow instead, and the scenario typically stops safely.

\begin{table}[!ht]
\centering\footnotesize
\caption{Fault operators, their target streams, and the parameter ranges explored by specification-space search.}
\label{tab:faults}
\begingroup
\setlength{\tabcolsep}{4pt}
\setlength{\aboverulesep}{0pt}
\setlength{\belowrulesep}{0pt}
\renewcommand{\arraystretch}{1.1}
\begin{tabularx}{\linewidth}{l l >{\raggedright\arraybackslash}X >{\raggedright\arraybackslash}p{2.7cm}}
\toprule
\textbf{Operator} & \textbf{Target} & \textbf{Effect} & \textbf{Search range} \\
\midrule
\rowcolor{TabGroup}\multicolumn{4}{l}{\textbf{Delivery faults}} \\
\hspace{0.5em}\code{drop} & any & Suppresses the next $k$ broadcasts in the window (all if $k$ is unset) & $k\in[1,40]$ \\
\hspace{0.5em}\code{burst\_drop} & any & Gilbert--Elliott burst loss with transitions $p_{\mathrm{gb}}, p_{\mathrm{bg}}$ and a bad-state loss rate & pathology library \\
\hspace{0.5em}\code{delay} & any & Constant delay plus jitter, optionally applied after reception (decode delay) & $[0.05,5]$\,s \\
\rowcolor{TabGroup}\multicolumn{4}{l}{\textbf{Timestamp fault}} \\
\hspace{0.5em}\code{skew} & any & Shifts the reported timestamp without delaying delivery, so the receiver misreads the age & $|\mathrm{skew}|\leq 30$\,s \\
\rowcolor{TabGroup}\multicolumn{4}{l}{\textbf{Semantic-content faults}} \\
\hspace{0.5em}\code{freeze} & SPaT, BSM & Holds \code{phase}, \code{countdown}, or \code{kinematics} at its value when the window opens & window only \\
\hspace{0.5em}\code{rebind} & MAP & Lanes of one movement report the signal group of another & movement pair \\
\hspace{0.5em}\code{pos\_perturb} & BSM & Position noise $\sigma$ plus a constant along-track bias $b$ & $\sigma\in[0.2,10]$\,m, $b\in[0,40]$\,m \\
\bottomrule
\end{tabularx}
\endgroup
\end{table}

\section{Safety Application Logic}
\label{app:apps}
Each application is a deterministic kinematic-threshold function $\pi_a(\beta,x)$ (Eq.~(\ref{eq:decision})). The same function drives both rollouts: the message-driven ego evaluates it on the reconstructed state $\beta$ and the ground-truth ego on the corresponding state derived from $W$, so any difference in their outcomes is attributable to the information channel. Braking decisions use a comfortable deceleration $a_c=3.5$\,m/s$^2$ and a reaction time $t_r=1.0$\,s, giving the stopping distance
\begin{equation}
d_s(v)=v\,t_r+\frac{v^2}{2a_c},
\label{eq:stopping}
\end{equation}
and the decision epoch of a signal application is the last point at which stopping remains comfortable, $d\leq d_s(v)$, where $d$ is the distance to the stop bar. Table~\ref{tab:apps} lists each decision rule and the violation that sets $V=1$ in Eq.~(\ref{eq:admission}).

Violations are adjudicated identically for both rollouts. A red entry is the crossing of the stop bar while the true signal of the ego's group is red; being beyond the bar when a later red begins is legal. A collision requires a center-to-center distance below the stated threshold while either vehicle moves faster than 0.5\,m/s. For FCW and EEBL, only vehicles ahead of the ego count, and the ego must be the striking vehicle. In CARLA, a vehicle counts toward a conflict only if it lies within 2.5\,m of the ego's route, so that adjacent-lane traffic passing close to the ego is not scored as a crash.

The monitor that identifies divergence episodes uses the budgets of Table~\ref{tab:budgets}. A signal belief is in error when its phase differs from the true phase or its countdown differs by more than 1.0\,s; a neighbor track is displaced when its position error exceeds the position budget plus its age multiplied by the vehicle's speed; and a believed track whose vehicle no longer exists becomes a ghost after 2.0\,s.

\begin{table}[!ht]
\centering\footnotesize
\caption{Decision rules and violation criteria of the six safety applications.}
\label{tab:apps}
\begingroup
\setlength{\tabcolsep}{4pt}
\setlength{\aboverulesep}{0pt}
\setlength{\belowrulesep}{0pt}
\renewcommand{\arraystretch}{1.1}
\begin{tabularx}{\linewidth}{l >{\raggedright\arraybackslash}X >{\raggedright\arraybackslash}p{2.5cm}}
\toprule
\textbf{App.} & \textbf{Decision rule} & \textbf{Violation} \\
\midrule
\rowcolor{TabGroup}\multicolumn{3}{l}{\textbf{Signal applications}} \\
\hspace{0.5em}\textbf{RLVW} & Predicts the phase at arrival; once $d\leq d_s(v)$, stops if arrival falls on red, or on yellow after the countdown expires. Holds on a current red within 6\,m of the bar; without signal data, stops at the decision point & Red entry \\
\hspace{0.5em}\textbf{GLOSA} & Keeps speed if arriving within the current green; otherwise advises $v=d/t_{\mathrm{green}}$, clipped to $[3,16]$\,m/s & Red entry \\
\rowcolor{TabGroup}\multicolumn{3}{l}{\textbf{Conflict applications}} \\
\hspace{0.5em}\textbf{IMA} & Stops if a crossing track reaches the conflict point within 3.5\,s of the ego (ego arrival within 12\,s) & Collision, $<1.8$\,m \\
\hspace{0.5em}\textbf{LTA} & Stops if an opposing track arrives before the ego's arrival plus a 4.0\,s gap (12\,s horizon) & Collision, $<1.8$\,m \\
\hspace{0.5em}\textbf{FCW} & Brakes if the time to collision with the lead is below 2.5\,s; holds behind a stopped lead within 10\,m & Forward collision, $<2.0$\,m \\
\hspace{0.5em}\textbf{EEBL} & Brakes on a relayed hard-braking flag from a lead vehicle within 80\,m & Forward collision, $<2.0$\,m \\
\bottomrule
\end{tabularx}
\endgroup
\end{table}

\section{Failure Margin and Search Settings}
\label{app:margin}
The failure margin $m(S)$ of Eq.~(\ref{eq:margin-objective}) measures how close an executed, non-admitted scenario came to failing; smaller is closer. For signal applications it is the smallest non-negative hazard slack over all divergence episodes,
\begin{equation}
m(S)=\min_{e:\,\delta_e\geq 0}\delta_e,
\qquad
\delta_e=\min_{t_d\in\mathcal{U}_e}\,(t_d-t^{\mathrm{end}}_e),
\label{eq:margin-signal}
\end{equation}
where $\mathcal{U}_e$ is the set of decision epochs $t_d$ of the message-driven rollout that occur after episode $e$ begins, and $t^{\mathrm{end}}_e$ is the episode's end. A small slack means the wrong belief ended just before a decision; a negative slack means a decision was taken while the belief was wrong. For conflict applications, $m(S)$ is the minimum distance between the message-driven ego and any agent over the run, counting only vehicles ahead of the ego for FCW and EEBL. If no episode has non-negative slack (including a run without episodes) or no conflict occurs, $m(S)$ is set to a large constant; a run whose episodes all have negative slack therefore keeps this constant.

Specification-space search starts from the compiled scenario and keeps an elite set of the $K=8$ highest-priority executed scenarios. Each iteration samples a parent with probability proportional to $w(S)$ (Eq.~(\ref{eq:search-weight})); a scenario in which both rollouts violate is non-attributable and receives the fixed priority 0.05 instead, the weight of a scenario with $m(S)=19.5$, so it remains a possible parent but ranks below every closer near miss. One of the operators in Table~\ref{tab:mutations} is drawn, with weights that favor interaction-timing operators for conflict applications and fault operators for signal applications. Mutations never change the application, ego maneuver, agent roster, or map query, so the test intent is preserved. Each mutant passes through the verifier before execution, and the budget of $N_s=120$ candidates per scenario counts both executed and gated mutants. Admitted scenarios remain in the elite set so that search can explore neighboring mechanism cells. Each admitted scenario is minimized by delta debugging (Section~\ref{sec:refine}). The refinement study runs search with a fixed random seed on the stub backend, and allows up to $N_r=5$ reflection rounds per scenario.

\begin{table}[!ht]
\centering\footnotesize
\caption{Mutation operators of specification-space search. $^{\ast}$Replaces the scenario's fault vector.}
\label{tab:mutations}
\begingroup
\setlength{\tabcolsep}{4pt}
\setlength{\aboverulesep}{0pt}
\setlength{\belowrulesep}{0pt}
\renewcommand{\arraystretch}{1.1}
\begin{tabularx}{\linewidth}{>{\raggedright\arraybackslash}p{3.1cm} >{\raggedright\arraybackslash}X}
\toprule
\textbf{Operator} & \textbf{Change} \\
\midrule
\rowcolor{TabGroup}\multicolumn{2}{l}{\textbf{Fault vector}} \\
\hspace{0.5em}Fault magnitude & Scales a delay or skew by $\times0.5$ or $\times2$; changes a drop count by $-5$ to $+10$; scales $\sigma$ or shifts $b$ \\
\hspace{0.5em}Fault window & Shifts the offset by $\pm2.5$\,s, resets the duration, or changes the anchor event \\
\hspace{0.5em}Stuck feed$^{\ast}$ & SPaT \code{phase} and \code{countdown} freeze opening 2.5 or 5\,s before the amber onset \\
\hspace{0.5em}Outage$^{\ast}$ & Window-long drop of the consumed stream (5, 8, or 12\,s) \\
\hspace{0.5em}Track freeze$^{\ast}$ & Freezes neighbors' BSM kinematics from 3--8\,s before the ego's decision point \\
\hspace{0.5em}Position bias$^{\ast}$ & Persistent BSM bias of 8, 15, or 25\,m ($\sigma=0.5$\,m) \\
\hspace{0.5em}Pathology swap$^{\ast}$ & Library pathology at a random intensity \\
\rowcolor{TabGroup}\multicolumn{2}{l}{\textbf{Interaction and equipage}} \\
\hspace{0.5em}Conflict alignment & Places a conflict agent so that it arrives with the ego, to within $\pm8\%$ \\
\hspace{0.5em}Agent kinematics & Shifts an agent's distance by 5--40\,m within $[5,300]$\,m, or its speed by 0.5--3\,m/s within $[0.5,35]$\,m/s \\
\hspace{0.5em}Ego speed & Shifts the approach speed by $\pm2$\,m/s within $[2,25]$\,m/s \\
\hspace{0.5em}Penetration & Shifts the market penetration rate by $\pm0.25$ within $[0,1]$ \\
\hspace{0.5em}SPaT rate & Halves or doubles the SPaT rate within $[1,20]$\,Hz \\
\bottomrule
\end{tabularx}
\endgroup
\end{table}

\section{LLM Prompts}
\label{app:prompts}
\begin{figure*}[!t]
\centering
\begin{minipage}{\textwidth}
\begin{promptpanel}{Parser}{Module I}
\begin{pfields}
\item[Role] Translate a natural-language V2X test intent into a scenario-specification JSON object for a CARLA stress-testing pipeline that evaluates SAE Day-1 safety applications under communication degradation. The six applications are listed with their descriptions and preconditions; choose exactly one.
\item[Rules] \begin{prules}
\item If the prompt describes only traffic, infer the application from the conflict (left turn with oncoming traffic $\rightarrow$ LTA; crossing vehicle $\rightarrow$ IMA; slower or stopped vehicle ahead $\rightarrow$ FCW; relayed hard braking $\rightarrow$ EEBL; signalized approach about arriving on green $\rightarrow$ GLOSA; other signalized approaches $\rightarrow$ RLVW) and assume a healthy, standards-nominal channel (SPaT 10\,Hz, MAP 1\,Hz, BSM 10\,Hz, full penetration, no faults); never invent a degradation.
\item Transcribe every vehicle mentioned, including background traffic.
\item Time each conflict so that the agent's distance divided by its speed matches the ego's arrival.
\item Express degradations only with the seven fault primitives and anchored windows; a pathology library describes known patterns in these terms.
\item A fault must be able to breach the application's age budget: a drop of $k$ messages at rate $f$ breaches budget $B_a$ only if $k/f>B_a$. A plain SPaT outage is usually fail-safe, so a stuck feed is expressed as a \code{freeze} of phase and countdown opened before the phase change.
\item Keep the 6/3/12\,s signal plan, the driver-character presets, the weather, and the channel defaults unless the prompt states otherwise.
\item Respond with only the JSON object.
\end{prules}
\item[Repair] A schema violation or blocking verifier finding is returned as a follow-up message (``That JSON failed validation: \ldots'' or ``The spec was rejected by the feasibility verifier: \ldots'') asking for a corrected object, for up to three retries.
\end{pfields}
\end{promptpanel}
\begin{promptpanel}{Planner}{Module II}
\begin{pfields}
\item[Role] Place the specified agents at a junction that is already bound, given live junction facts: the ego's approach, the fixed signal cycle, and the candidate approaches for each conflict role with their true lane headings.
\item[Rules] \begin{prules}
\item Keep the same agent identifiers and do not add or remove agents.
\item Assign crossing and oncoming agents to one of their role's candidate approaches; agents ahead, behind, or in the background share the ego's lane.
\item Time each conflict so that the agent arrives within a few seconds of the ego.
\item Respect each agent's character: stopped vehicles stay stopped, aggressive agents may run the red, and other agents comply with their signal.
\end{prules}
\item[Output] A JSON plan with a rationale, the ego approach speed, and per-agent approach, distance, speed, and behavior; invalid plans are returned with the problems listed.
\end{pfields}
\end{promptpanel}
\begin{promptpanel}{Reflection}{Module IV}
\begin{pfields}
\item[Role] Act as the refinement analyst. A prompt was compiled and executed as a paired rollout; a scenario is admitted only if the message-driven ego violates its safety criterion while the ground-truth ego does not. Given the prompt, the evidence digest, and a goal, diagnose why the goal was not met and revise the prompt.
\item[Evidence digest] Admission status; each rollout's violation and minimum conflict gap; the active faults; divergence episodes with onset, end, hazard slack, and whether they are structural; AoI at the decision epochs; and verifier findings.
\item[Rules] \begin{prules}
\item Keep the scenario's identity (place, conflict story, application) and change only timing, distances, speeds, fault strength or anchoring, equipage, or wording the parser misread.
\item Write plain English, not JSON or parameter names.
\item Change as little as plausibly moves the run toward the goal, and escalate decisively if an earlier revision changed nothing.
\end{prules}
\item[Domain notes] \begin{pnotes}
\item Delivery faults make the belief stale, which the age budget detects, so applications usually fall back safely; content faults keep a wrong belief looking fresh.
\item A frozen signal matters only if it masks a phase change and persists through the decision.
\item A conflict matters only if the arrivals coincide and the vehicle is within V2V range early enough.
\item A persistent position bias is more potent than noise of the same size.
\item Describing a feed as ``stuck'' compiles to a persistent hold, whereas ``freezes for a few seconds'' compiles to a short one.
\end{pnotes}
\item[Output] A JSON object with a diagnosis citing the evidence, the list of changes, and the full revised prompt, which re-enters Module~I.
\end{pfields}
\end{promptpanel}
\end{minipage}
\caption{Condensed system prompts of the three LLM roles of \sys{}.}
\label{fig:prompts}
\end{figure*}

\sys{} uses three LLM roles that share one backend, GPT-4o at temperature 0.2 with a fixed sampling seed (Section~\ref{sec:experiments}). Each receives a fixed system prompt, condensed in Fig.~\ref{fig:prompts}, and must return a single JSON object that is validated before use. Rules are paraphrased, and the worked examples in the parser prompt are omitted. The domain notes given to the reflection role come from the V2X knowledge base and are identical for every scenario; they are not tuned to individual prompts.

\section{Reproducibility}
\label{app:repro}
Table~\ref{tab:experiments} summarizes the configuration of each experiment. All runs are deterministic given a scenario seed, which fixes vehicle equipage, message-bus jitter, fault stochasticity, RSU pole selection, vehicle blueprints, and retrieval tie-breaking, so every scene replays identically. Each admitted failure is stored with its full minimal-witness specification and re-verified against the paired admission test on replay. For every benchmark scene, \sys{} archives the compiled specification, the scene plan with its rationale, the grounding record, the execution trace, the monitor report, three camera streams with rendered V2I/V2V links, and RSSI, latency, and delivery maps. Further material is available on the project page: \url{https://keshuw95.github.io/LAST-V2X/}.

\begin{table}[!htb]
\centering\footnotesize
\caption{Experiment configurations. All runs use GPT-4o (temperature 0.2, fixed sampling seed) and synchronous live CARLA simulation, except $^{\ddagger}$search runs, which use the stub backend; the headline mechanisms were validated in live CARLA.}
\label{tab:experiments}
\begingroup
\setlength{\tabcolsep}{4pt}
\setlength{\aboverulesep}{0pt}
\setlength{\belowrulesep}{0pt}
\renewcommand{\arraystretch}{1.1}
\begin{tabularx}{\linewidth}{>{\raggedright\arraybackslash}p{3.6cm} >{\raggedright\arraybackslash}X l}
\toprule
\textbf{Experiment} & \textbf{Scale} & \textbf{Seeds} \\
\midrule
Normal-traffic benchmark & 100 prompts, 8 towns & 1 per scene \\
Communication-fault suite & 12 prompts & 1 per scene \\
Corridor traversal & 6 prompts, Town05; 3 offset plans $\times$ 3 radio ranges & 10 (420 runs) \\
Multiple connected vehicles & 6 prompts, Town05 & 10 per condition \\
Refinement suite$^{\ddagger}$ & 50 prompts; $\leq$120 search candidates and $\leq$5 reflection rounds per scene & 1 per scene \\
\bottomrule
\end{tabularx}
\endgroup
\end{table}

\end{document}